\documentclass[times]{TRR}

\usepackage{etoolbox}
\makeatletter
\AtBeginDocument{%
\@ifpackageloaded{natbib}{%
\citestyle{authoryear}%
\setcitestyle{authoryear,round,semicolon,aysep={,},yysep={;}}%
\renewcommand\@biblabel[1]{}%
\renewcommand{\bibnumfmt}[1]{}%
}{}%
}
\makeatother

\usepackage{moreverb,url}

\usepackage[colorlinks,bookmarksopen,bookmarksnumbered,citecolor=red,urlcolor=red]{hyperref}

\newcommand\BibTeX{{\rmfamily B\kern-.05em \textsc{i\kern-.025em b}\kern-.08em
T\kern-.1667em\lower.7ex\hbox{E}\kern-.125emX}}

\begin{document}

\runninghead{Bjelonic et al.}

\title{Towards bridging the gap: Systematic sim-to-real transfer for diverse legged robots}

\author{Filip Bjelonic\affilnum{1}, Fabian Tischhauser\affilnum{1} and Marco Hutter \affilnum{1}}

\affiliation{\affilnum{1}Robotic Systems Lab, ETH Zurich, Zurich, Switzerland\\
}

\corrauth{Filip Bjelonic, filip@bjelonic.com}

\begin{abstract}
Legged robots must achieve both robust locomotion and energy efficiency to be practical in real-world environments. Yet controllers trained in simulation often fail to transfer reliably, and most existing approaches neglect actuator-specific energy losses or depend on complex, hand-tuned reward formulations.

We propose a framework that integrates sim-to-real reinforcement learning with a physics-grounded energy model for permanent magnet synchronous motors. The framework requires a minimal parameter set to capture the simulation–reality gap and employs a compact four-term reward with a first-principle-based energetic loss formulation that balances electrical and mechanical dissipation. 

We evaluate and validate the approach through a bottom-up dynamic parameter identification study, spanning actuators, full-robot \emph{in-air} trajectories and \emph{on-ground} locomotion. The framework is tested on three primary platforms and deployed on ten additional robots, demonstrating reliable policy transfer without randomization of dynamic parameters. Our method improves the energetic efficiency over state-of-the-art methods, achieving a \qty{32}{\percent} reduction in the full Cost of Transport of \robot{ANYmal} (\num{1.27}).
All code, models, and datasets are publicly available.
\end{abstract}

\keywords{Legged robots, Quadrupedal locomotion, Reinforcement learning, Simulation-to-real transfer, Reality gap, Energy efficiency}

\maketitle

\begin{figure}
    \centering

    \begin{subfigure}{0.49\linewidth}
        \centering
        \includegraphics[width=\linewidth]{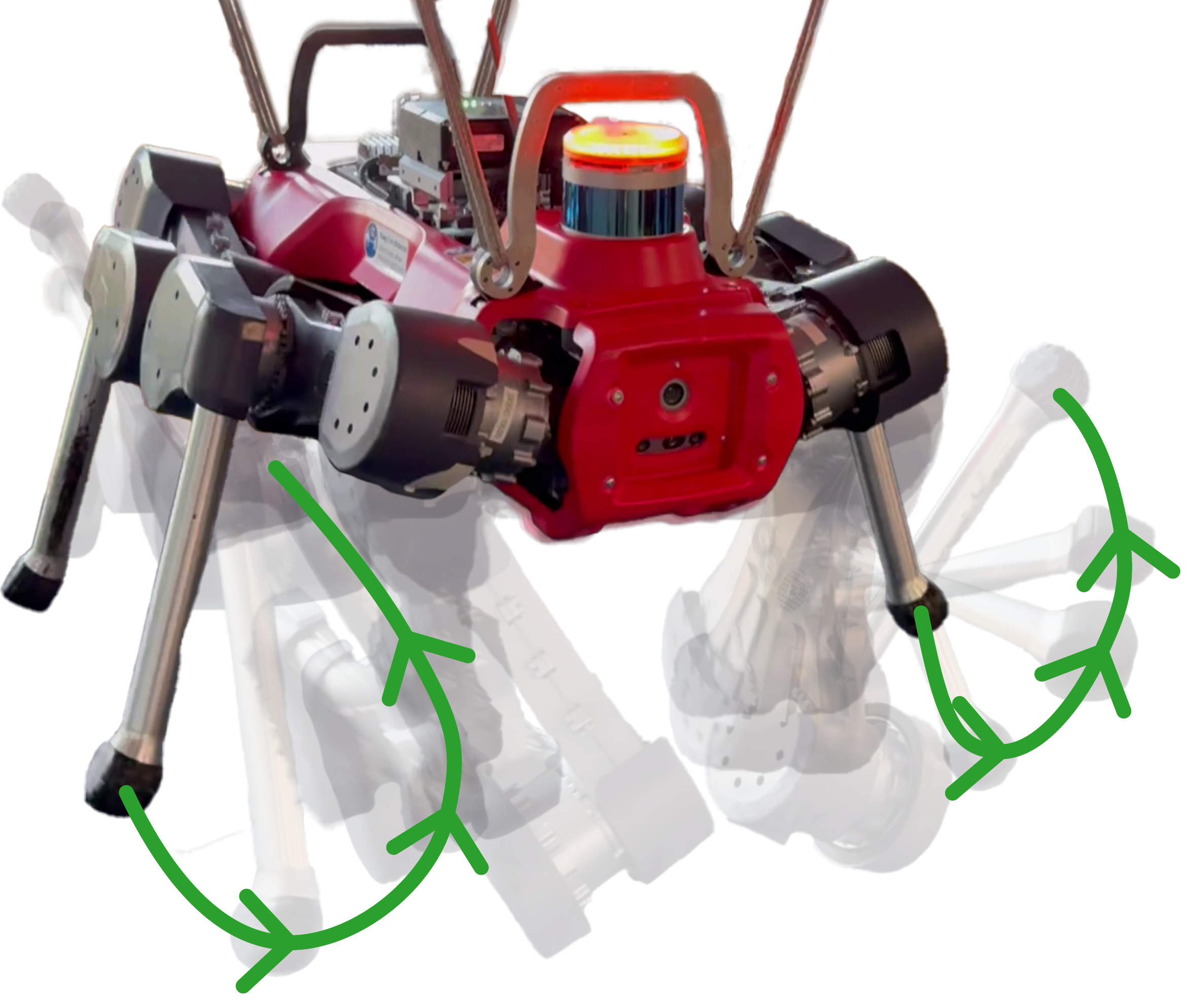}
        \caption{Real}
        \label{fig:figure1:real_stills}
    \end{subfigure}\hfill
    \begin{subfigure}{0.49\linewidth}
        \centering
        \includegraphics[width=\linewidth]{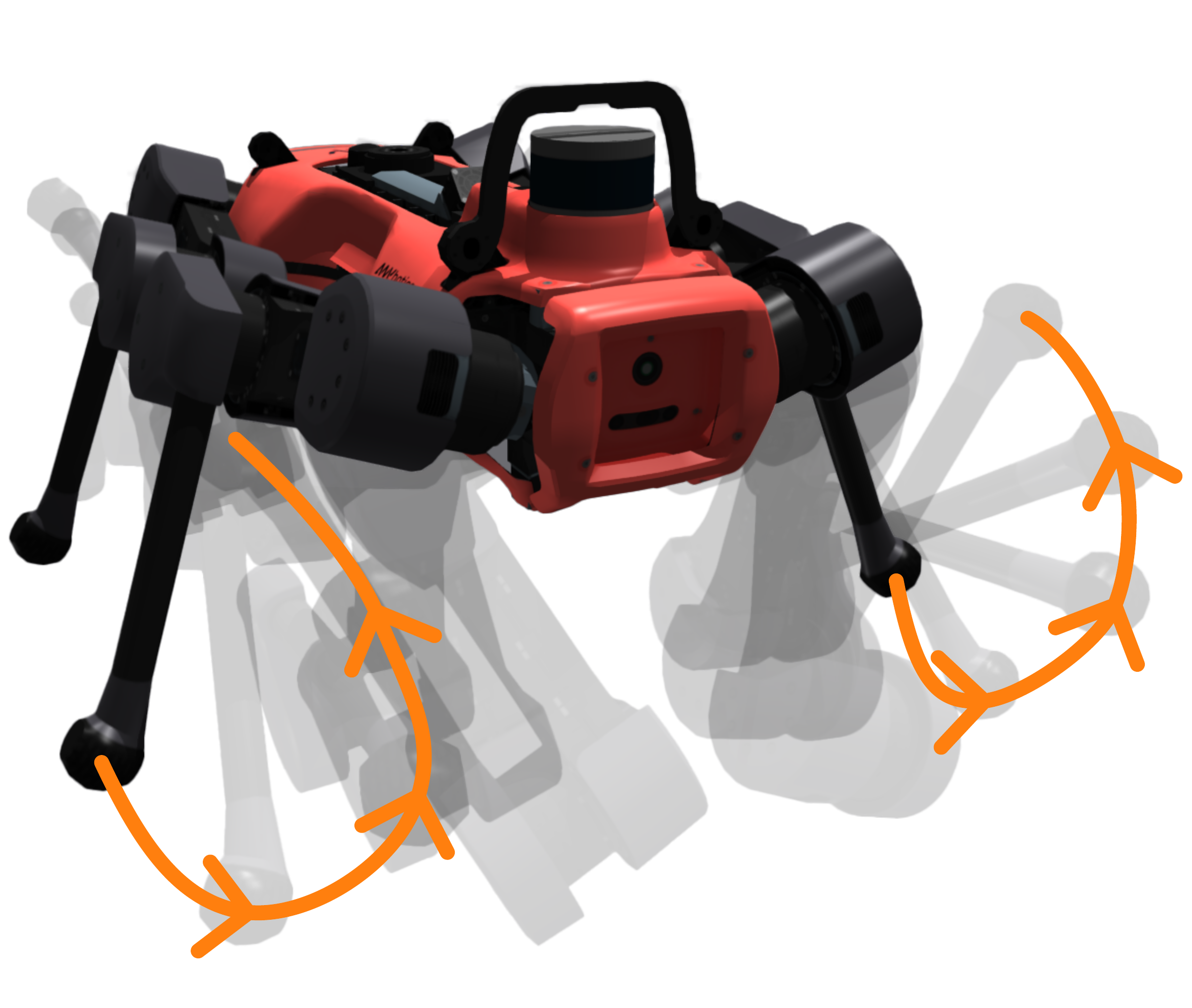}
        \caption{Sim (ours)}
        \label{fig:figure1:sim_stills}
    \end{subfigure}

    \begin{subfigure}{\linewidth}
        \centering
        \includegraphics[width=\linewidth]{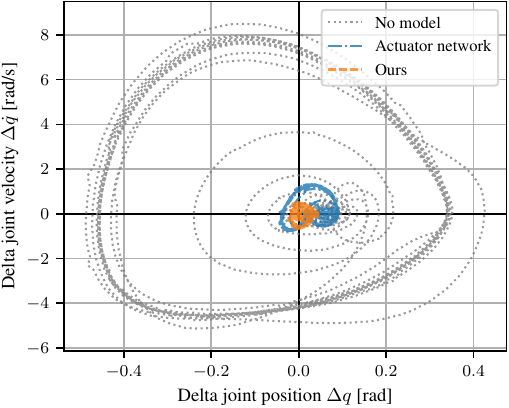}
        \caption{Delta phase portrait}
        \label{fig:figure1:phase_portrait}
    \end{subfigure}

    \caption{Comparison of real and simulated robot trajectories under a 0.1 – \SI{2.0}{\hertz} chirp input. (a) Picture of the real robot with an overlaid trace showing its motion sequence from experiment. (b) Equivalent view in simulation using our modeling approach. In both (a) and (b), the robot’s base is suspended so that its legs swing freely. (c) Phase portrait of the “delta” joint state—i.e., at each timestep, the real joint position minus the simulated joint position (x-axis) and the real joint velocity minus the simulated joint velocity (y-axis)—plotted separately for three modeling approaches: no model (light gray), state-of-the-art actuator network (blue), and our proposed method (orange). Points nearer to (0, 0) indicate that the simulated trajectory more closely matches the real trajectory.}
    \label{fig:main_figure}
\end{figure}

\noindent Legged robots promise versatile mobility in environments that are inaccessible to wheeled or tracked platforms. However, achieving robust and efficient locomotion remains a central challenge: Controllers trained in simulation only show  the same performance on the real system if the system and actuator dynamics are accurately modeled. Modeling errors, in particular in the actuator dynamics, where electrical and mechanical losses are often not properly modeled, lead to major inefficiencies and hence directly affect robot endurance and payload capacity.

Prior work has addressed the simulation-to-reality gap through extensive domain randomization, residual modeling, or full physics identification. While these methods have demonstrated impressive sim-to-real transfer, they often require specialized sensors, exhaustive parameter identification, or iterative expert tuning of heuristic models and reward functions. As a result, current RL controller pipelines often rely on complex, high-dimensional reward functions. As modeling errors increase the sim-to-real gap, environment design often resorts to repeatedly adjusting rewards based on observed discrepancies in the real system -- an ad-hoc process that can be repeated many times without clear structure, and which further disconnects rewards from physical notions such as efficiency.

This work introduces a framework (Figure~\ref{fig:approach_overview}) that addresses these challenges by combining a sim-to-real RL pipeline with systematic actuator modeling, with its core modeling advantage illustrated in Figure~\ref{fig:main_figure}. We perform a bottom-up analysis: first characterizing single actuator models, then estimating dynamic parameters on the full-system level, and finally validating locomotion control across multiple platforms.
\rev{A focus of our framework is the integration of a physics-grounded energy model for \acp{pmsm}, which enables training of locomotion policies that are both transferable and energy-efficient.
This work targets the reduction of \emph{robot-specific reward tuning} for locomotion by grounding the dominant regularization term -- energy consumption -- in physical units that transfer across morphologies. The resulting locomotion objective consists of four independently weighted terms: velocity tracking, energy consumption, collision avoidance, and foot-touchdown velocity. The first two encode task performance and physical efficiency, while the latter two act as structural safety terms that discourage damaging contacts and impact-heavy gaits.}

The main contributions of this work are:
\begin{enumerate}
    \item \textbf{Fixed-base excitation driven system identification:} A data-efficient identification pipeline based exclusively on fixed-base joint excitation, enabling systematic actuator-level calibration without relying on locomotion rollouts or motion priors. (see Section~\ref{sec:method}).
    \item \textbf{Bottom-up performance analysis:} Multi-level evaluation from single actuators to full-robot locomotion, including comparisons with state-of-the-art black-box approaches (see Section~\ref{sec:experiments:model_analysis}).
    \item \textbf{Cross-platform validation:} Deployment on three primary platforms (\robot{ANYmal}, \robot{Tytan}, \robot{Minimal}) and ten additional systems (see Section~\ref{sec:experiments:sim2real}).
    \item \textbf{Energetic assessment:} Quantitative analysis of electrical and mechanical efficiency, demonstrating improvements over previous methods (see Section~\ref{sec:experiments:running_track}).
\end{enumerate}

Additional contributions include:
\begin{itemize}
    \item A joint position control strategy with position saturation for hardware protection during locomotion (see Section~\ref{sec:method:safe_pd_control}).
    \item Empirical evidence that four reward terms suffice for effective locomotion training (see Section~\ref{sec:method:rewards} and \ref{sec:results:sim2real}).
    \item A physics-grounded energetic reward that balances electrical and mechanical losses to minimize overall consumption (see Section~\ref{sec:method:rewards}).
\end{itemize}

All code, models, and raw data underlying the results, including those for most figures and tables, are publicly available.

\section{Related Work}
\begin{figure}
    \centering

    \begin{subfigure}{0.4\linewidth}
        \centering
        \includegraphics[width=\linewidth]{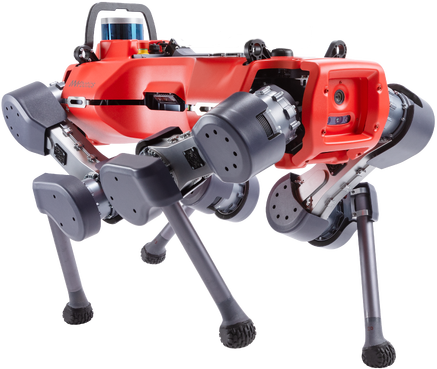}
        \caption{\robot{ANYmal}~D}
    \end{subfigure}\hfill
    \begin{subfigure}{0.4\linewidth}
        \centering
        \includegraphics[width=\linewidth]{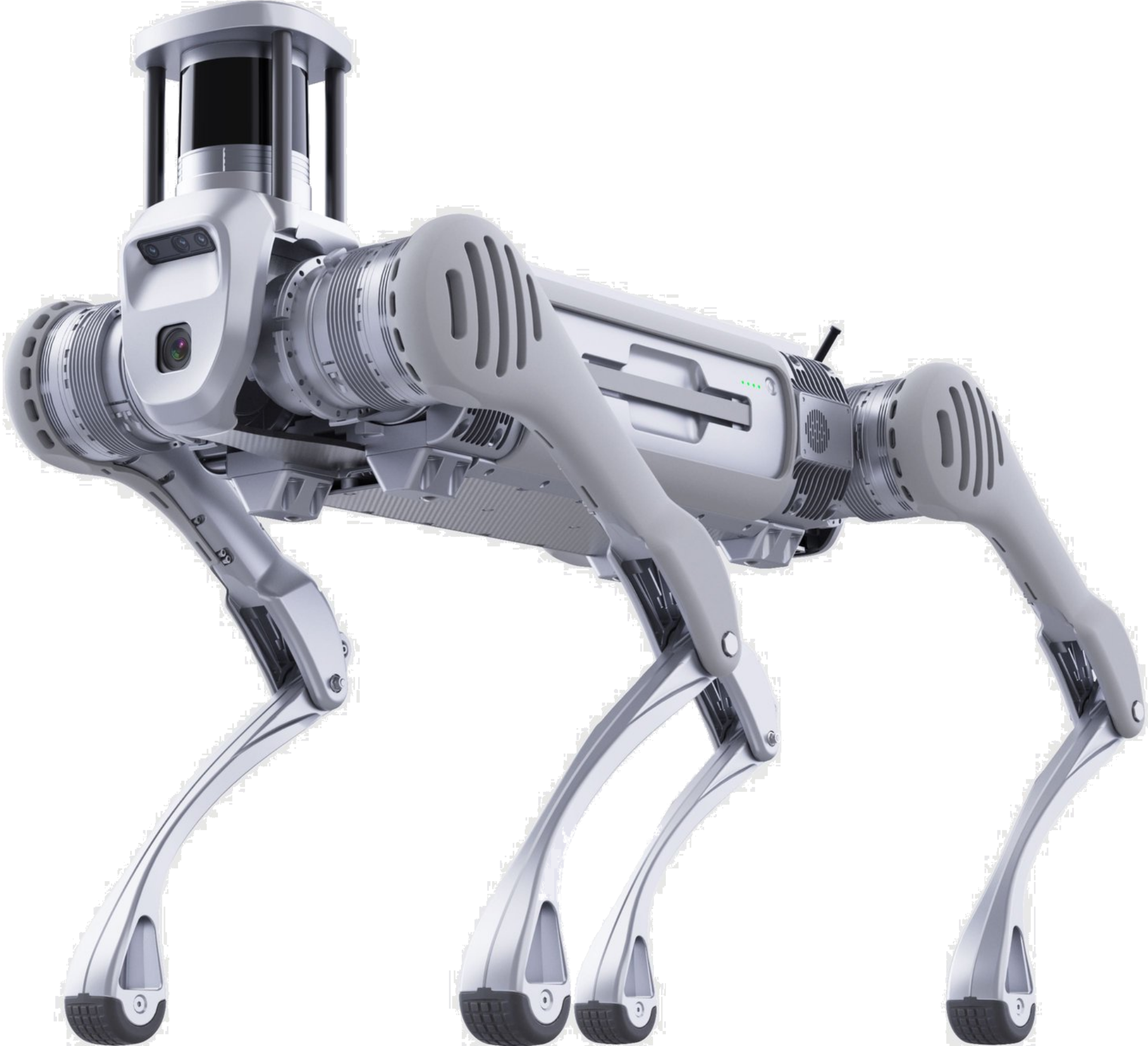}
        \caption{\robot{B2}}
    \end{subfigure}

    \medskip

    \begin{subfigure}{0.3\linewidth}
        \centering
        \includegraphics[width=\linewidth]{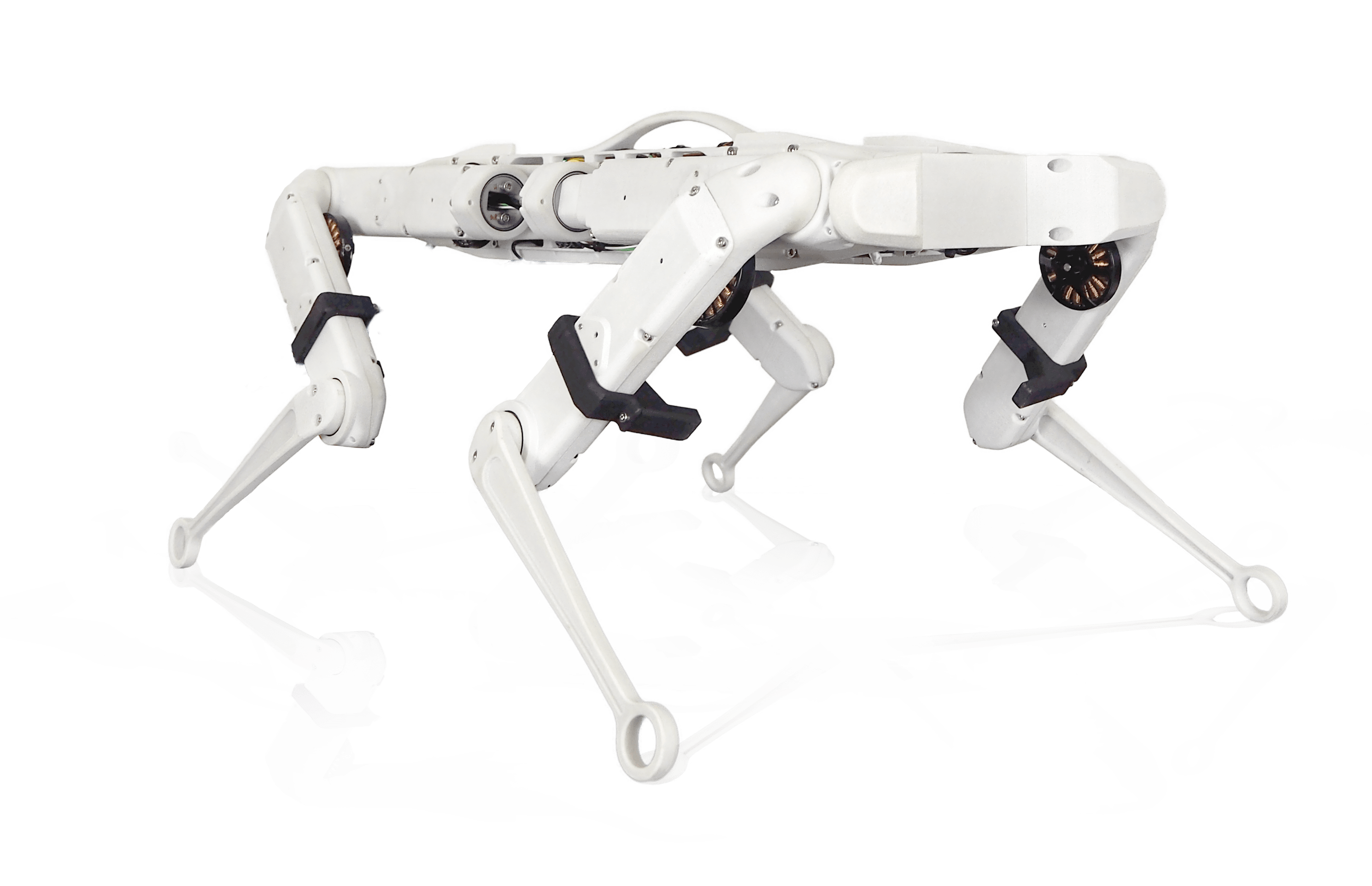}
        \caption{\robot{Solo}}
    \end{subfigure}\hfill
    \begin{subfigure}{0.3\linewidth}
        \centering
        \includegraphics[width=\linewidth]{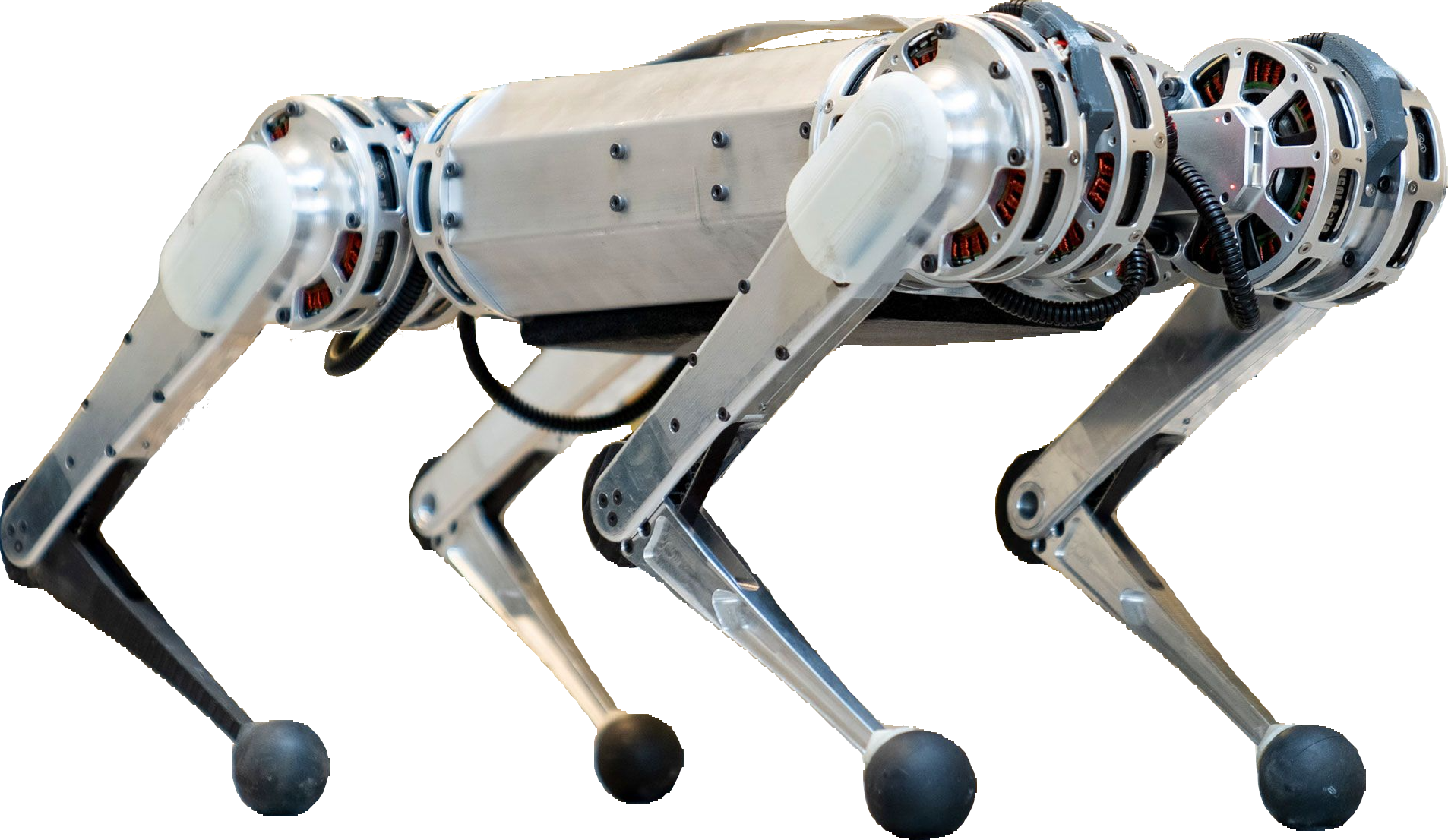}
        \caption{\robot{Mini-Cheetah}}
    \end{subfigure}\hfill
    \begin{subfigure}{0.2\linewidth}
        \centering
        \includegraphics[width=\linewidth]{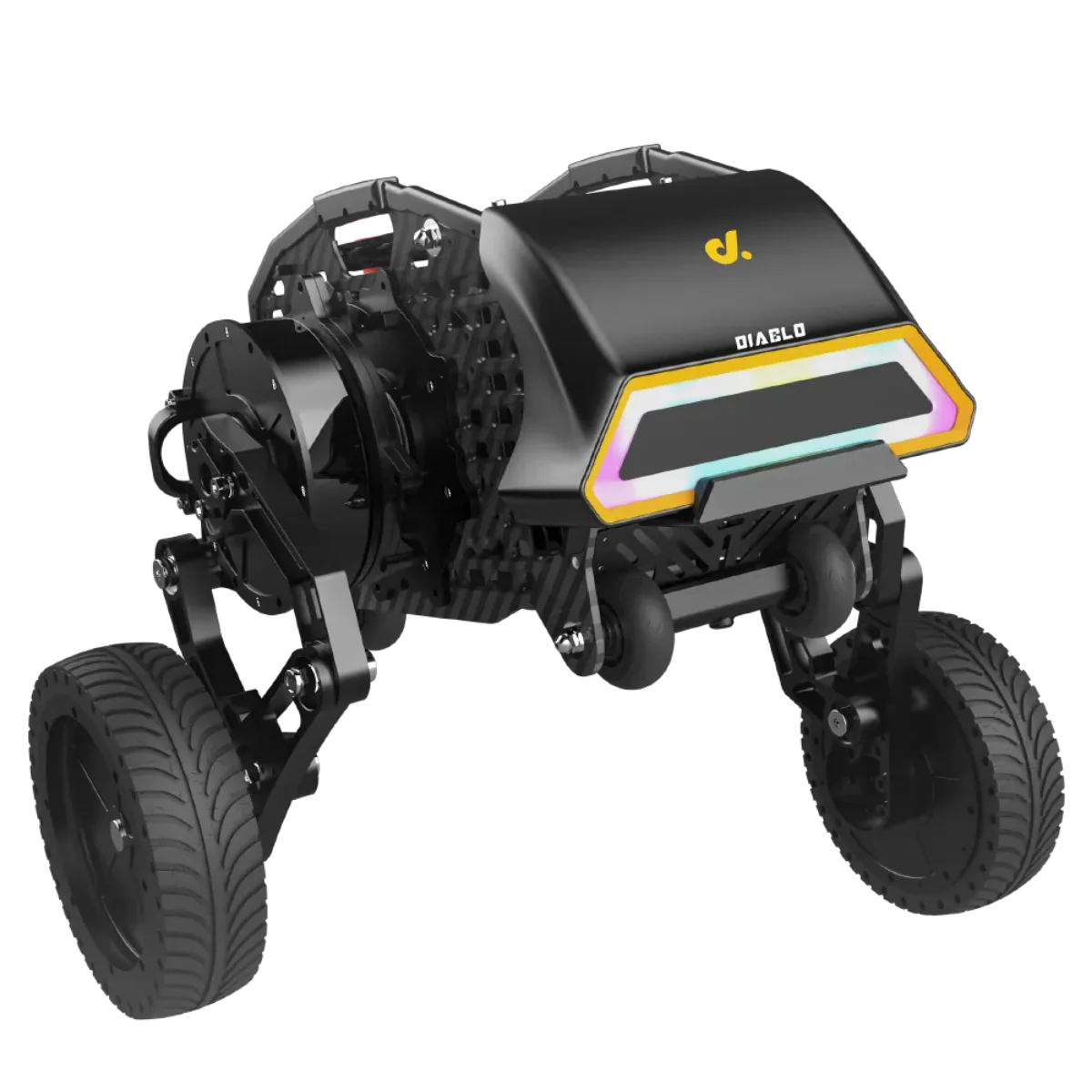}
        \caption{\robot{DIABLO}}
    \end{subfigure}

    \caption{Representative legged robots with explicitly documented use of \ac{pmsm} for actuation.}
    \label{fig:3x3grid}
\end{figure}

Our manuscript focuses on state-of-the-art methods for rigid robots actuated by \acfp{pmsm}, which are the dominant choice in modern legged robots due to their efficiency and controllability (see Figure~\ref{fig:3x3grid} for a few prominent examples~\cite{hutter2017anymal, unitree2025motorsdk, aractingi2023controlling, katz2018low, liu2024diablo}). For clarity, we note that while “\acf{bldc}” is often used in industry as a generic label for brushless motors, it formally refers to machines with trapezoidal back-EMF driven by six-step commutation, whereas most legged robots employ sinusoidal \acp{pmsm} with field-oriented control~\cite{derammelaere2016quantitative, gamazo2010position}. 

\subsection{Parameter estimation}
Parameter estimation for robotic systems has been extensively studied, with classical approaches relying on explicit dynamic models that are linear in the unknown parameters.
In such settings, weighted least-squares, maximum-likelihood estimation~\cite{swevers2002optimal}, or Kalman filtering~\cite{gautier2001extended} can be applied efficiently, provided that sufficiently exciting trajectories are available and that measurements of joint torques, velocities, and accelerations are reliable~\cite{kozlowski2012modelling}.
Comprehensive overviews of these approaches can be found in the system identification literature~\cite{wu2010overview}.

In practice, however, actuator-level dynamics often violate the assumptions required for closed-form estimation.
Nonlinear effects such as friction, saturation (cf. Figure~\ref{fig:torque_velocity_saturation_model}) and delays introduce discontinuities that render the loss function nonconvex and non-differentiable.
Moreover, access to accurate torque sensing is frequently unavailable or unreliable on legged platforms.

Bayesian optimization provides a probabilistic alternative that is effective for very low-dimensional problems with expensive evaluations~\cite{frazier2018bayesian}.
However, its performance typically degrades with increasing dimensionality and high-variance trajectory-level objectives, making it less suitable for moderate-dimensional parameter spaces~\cite{stork2022new} considered here.

Evolutionary strategies offer a robust alternative in this regime~\cite{beyer2007robust}.
They require minimal assumptions on smoothness, tolerate discontinuities, and parallelize naturally across large batches of simulations~\cite{stork2022new}.
Given the availability of massively parallel simulation~\cite{rudin2022learning}, we employ evolutionary strategies.
We emphasize that the proposed framework is agnostic to the specific choice of optimizer; \ac{cmaes} is used as a practical instantiation rather than a core contribution of this work.

\subsection{Trajectory comparison metrics.}
Prior sim-to-real and system identification work evaluates simulator quality using a range of trajectory similarity measures.
Common choices include time-domain mean-squared error in joint space (e.g. joint torques)~\cite{hwangbo2019learning},
frequency-domain metrics such as bandwidth agreement~\cite{kashiri2017sensor},
phase-space or limit-cycle comparisons for periodic motions using auto-correlation~\cite{albu2020review},
and task-level metrics such as tracking error or stability margins during deployment~\cite{misenti2025experimental}.
More advanced approaches employ distributional distances over trajectory segments~\cite{miller2025high} using Wasserstein distance~\cite{ruschendorf1985wasserstein}, Maximum Mean Discrepancy~\cite{dziugaite2015training} or identifiability-driven measures~\cite{sobanbabu2025sampling}.

In this work, we focus on direct joint-position trajectory matching under controlled in-air excitation. Because joint targets are replayed open-loop at the position-command level and contacts are removed, phase drift and long-horizon accumulation effects are negligible. Under these conditions, time-averaged joint position error provides a physically interpretable objective for identifying dominant actuator and transmission dynamics.

\begin{table*}
\small\sf\centering
\caption{Comparison of our framework with closely related sim-to-real calibration pipelines.}
\label{tab:related_work:pace_comparison}
\begin{tabular}{c|c|c|c}
\toprule
\textbf{Aspect} & \cite{miller2025high} & \cite{sobanbabu2025sampling} & Ours (PACE) \\
\hline \hline
Hardware data source 
& Locomotion policy rollouts
& Policies / motion priors
& Fixed-base joint excitation \\

Closed-loop control required 
& Yes 
& Yes 
& No \\

Safety during data collection 
& Medium (falls possible)
& Medium
& High (base fixed in-air) \\

Need for stable pre-trained policy
& Yes 
& Yes 
& No \\

Trajectory drift
& Yes 
& Yes 
& No \\

Primary optimized parameters 
& Friction, torque-speed curve 
& System-level parameters 
& Inertia, damping, friction, delay, bias \\

Parameters pre-computed 
& Delay 
& -- 
& Torque-speed curve \\

Optimization method 
& CMA-ES 
& CMA-ES 
& CMA-ES \\

Similarity metric 
& Wasserstein, MMD 
& Fisher information reward 
& Joint position error \\

Reason for metric choice 
& Drift in rollouts 
& Exploration reward 
& Simplicity (no drift) \\

Number of robots 
& 1 (Spot)
& 2 (Unitree focused)
& 3+ (3 main + 10 additional) \\
\bottomrule
\end{tabular}
\end{table*}

\subsection{Modeling}

We organize prior sim-to-real modeling approaches by the amount and role of real-world data used to reduce the reality gap. Along this spectrum, methods differ in how strongly they rely on physical priors versus learned components, and in whether data is used to induce robustness, correct residual effects, or directly learn predictive dynamics.

At one end, domain randomization requires little real-world data but assumes accurate rigid-body dynamics. Residual physics approaches combine moderate priors with learned corrections, needing additional data to capture unmodeled effects. At the other end, full dynamics models minimize assumptions but demand extensive real-world interaction to learn complete system dynamics. Both model-based and purely data-driven approaches are represented across this continuum.

\textbf{Low-data approaches.}
These methods primarily mitigate the sim-to-real gap by relying on strong physics priors from the robot description and robustness through domain diversity, rather than explicit system identification. Examples include Cassie and BDX droids SysID combined with domain randomization~\cite{xie2020learning, grandia2025design}, hand-calibrated parameters with randomized variations~\cite{li2024learning, tan2018sim}. Classical dynamics randomization for reinforcement learning~\cite{bellegarda2022robust} or domain adaptation for imitation learning~\cite{peng2020learning} fall into this category, where sim-to-real transfer is achieved through robustness induced by diverse simulated motion distributions rather than explicit dynamics alignment.
Large-scale randomization has been shown effective for dexterous manipulation (DexTreme~\cite{handa2023dextreme}). Variants include active domain randomization~\cite{mehta2020active}, probabilistic approaches such as BayesSim~\cite{ramos2019bayessim}, adversarial domain randomization~\cite{shi2024rethinking}, and methods like DROPO~\cite{tiboni2023dropo}.

\textbf{Moderate-data approaches.}
Residual models combine physics priors with learned corrections, often focusing on actuator or body-level dynamics. Neural augmentations of simulators capture uncertainties~\cite{ajay2018augmenting}, while adversarial techniques such as SimGAN identify simulation parameters~\cite{jiang2021simgan}. Neural-Augmented Simulation (NAS) introduces recurrent residual dynamics~\cite{golemo2018sim}.  
Joint-centric methods include actuator networks that leverage joint torque sensors~\cite{hwangbo2019learning}, unsupervised actuator models without torque sensing~\cite{fey2025bridging}, drive-limit estimations on \robot{Spot}~\cite{miller2025high}, and delta-action models through real-world rollouts (ASAP~\cite{he2025asap}).  
Body-centric methods aim to identify physical parameters at the system level, for instance via sampling-based active exploration (SPI-Active~\cite{sobanbabu2025sampling}) or learning discrepancies from human demonstrations (DROID~\cite{tsai2021droid}).  
Hybrid strategies combine low-level SysID with residual dynamics models, such as aerodynamic compensation for the floating robot \robot{BALLU}~\cite{sontakke2023residual}.

\textbf{High-data approaches.}
Full dynamics-based methods attempt to learn the complete robot dynamics with little or no reliance on physical priors. DayDreamer~\cite{wu2023daydreamer} learns world models from scratch, updating them through real-world rollouts while replaying the best policy. Offline world models similarly combine simulation-initialized dynamics with real-world fine-tuning~\cite{li2025offline}. Other approaches estimate grounded forward and inverse dynamics transformations~\cite{hanna2021grounded} to enhance simulator accuracy.

\textbf{Online adaptation.}
To cope with discrepancies that remain during deployment, online adaptation mechanisms are employed. Strategies include online fine-tuning~\cite{smith2022legged}, meta-learning for rapid adaptation~\cite{song2020rapidly}, and student–teacher schemes for online parameter identification~\cite{lee2020learning}. Recent approaches address shifts in dynamics during deployment by relying on the policy to implicitly infer or adapt to the underlying system at runtime, and can in principle recover the true system dynamics through online identification or adaptation~\cite{radosavovic2024real,li2025reinforcement}.

For comprehensive surveys on sim-to-real, see ~\cite{muratore2022robot, ju2022transferring}.

While prior work has successfully employed domain randomization~\cite{xie2020learning,handa2023dextreme} or residual dynamics~\cite{hwangbo2019learning,fey2025bridging}, often relying on implicit online adaptation, these approaches require specialized sensors or exhaustive parameter searches. Related actuator formulations have been used in prior work (e.g., Disney Research~\cite{grandia2025design}; Google~\cite{tan2018sim}), but without a systematic or validated procedure for reliably identifying their parameters on real hardware. Our method instead adopts a white-box actuator and joint dynamics model with partial observability, identifying a minimal and physically interpretable parameter set from encoder-only data to compactly capture the simulation–reality gap. We further show that actuator drive dynamics are largely linear ($H_q:\hat{q} \rightarrow q$), which enables fast optimization of a concise and physically interpretable parameter set that transfers across platforms without learned residuals.

We note that inertial parameters in the URDF are commonly obtained from CAD models or manufacturer specifications, and, when necessary, refined via experimental identification, as widely studied in classical robot dynamics and system identification literature~\cite{grandia2025design, gautier1986identification}.

\begin{figure*}
    \centering
    \includegraphics[width=\linewidth]{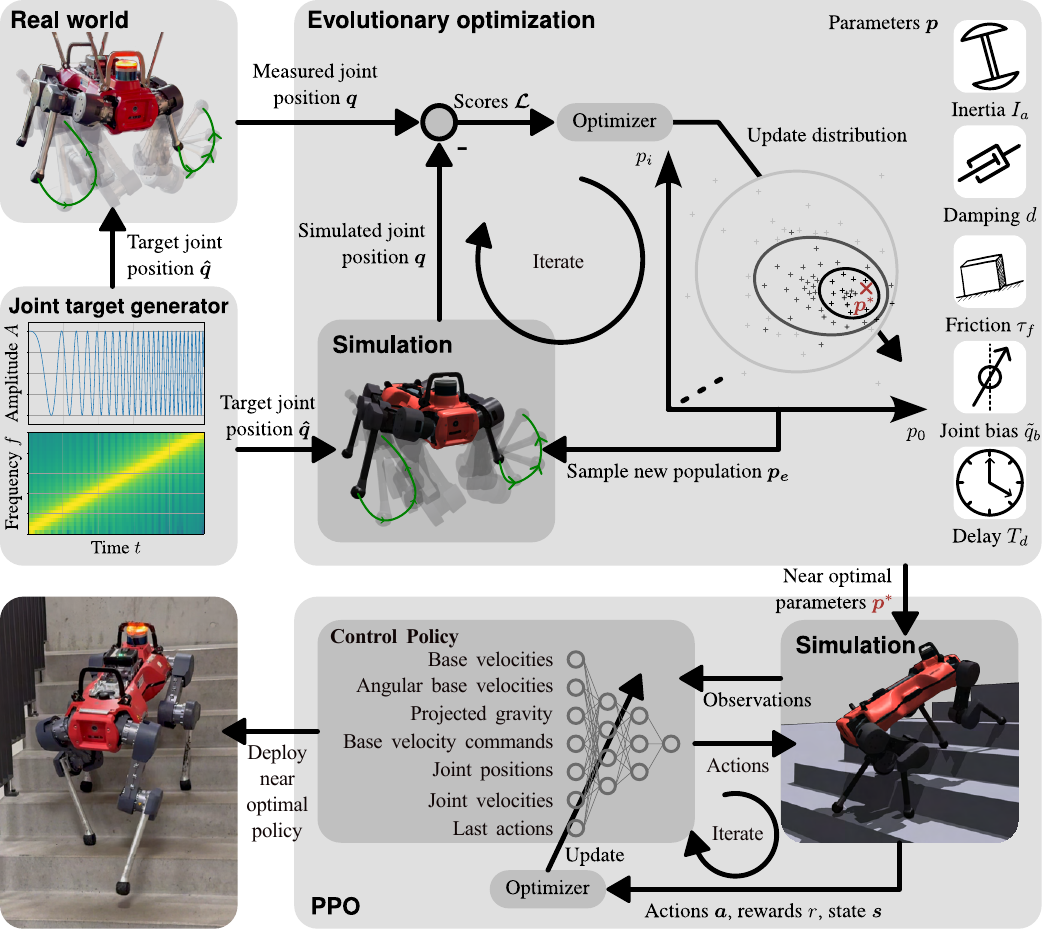}
    \caption{Overview of the proposed PACE pipeline with policy training. (i) Collection of real in-air data on a fixed-base setup (top left). (ii) Evolutionary parameter fitting of joint dynamics to align simulated and measured trajectories (top right). (iii) Blind policy training in simulation with zero-shot deployment on hardware (bottom).}
    \label{fig:approach_overview}
\end{figure*}

\subsubsection{Relation to policy-driven sim-to-real calibration.}

Several recent works propose iterative sim-to-real calibration pipelines that (i) collect hardware trajectories, (ii) optimize simulation parameters (e.g., via \ac{cmaes}), and (iii) train RL policies in the calibrated simulator for zero-shot deployment. In particular, ~\cite{miller2025high} and SPI-Active~\cite{sobanbabu2025sampling} follow this high-level structure using data collected from locomotion policies or motion priors on hardware and distributional or information-based objectives to match simulated and real behaviors. At this level, our framework shares the same overarching structure -- improving sim-to-real transfer by calibrating physically meaningful parameters -- and we therefore view these works as close references. We note that these works were developed largely concurrently with ours, reflecting parallel research directions in the community. A structured comparison between these approaches and ours is summarized in Table~\ref{tab:related_work:pace_comparison}. In this sense,~\cite{miller2025high} can be interpreted as a complementary extension, exploring how policy-driven data collection and distributional objectives can further refine sim-to-real transfer once a calibrated model is available.

Our method differs in the calibration signal and the resulting optimization regime: rather than relying on closed-loop locomotion rollouts (where errors compound, trajectories drift, and the collected data depends strongly on the current policy), we use repeatable fixed-base joint excitation to isolate actuator and joint dynamics under controlled conditions. This choice enables safe, low-cost data collection without requiring a stable locomotion controller, reduces confounding effects between control and dynamics, and allows direct trajectory-level error measures without suffering from long-horizon drift that motivates distributional matching objectives. Consequently, this method targets actuator-centric parameter observability and cross-platform applicability.

\subsection{Control}

Control of legged robots in complex terrain has been approached with both Model Predictive Control (MPC) and \ac{rl}~\cite{grandia2023perceptive,miki2022learning,xue2024full,rudin2025parkour}. Current state-of-the-art emphasizes \ac{rl}, often based on \ac{ppo}~\cite{schulman2017proximal} and its constrained derivatives, such as IPO~\cite{liu2020ipo} or TRPO-based formulations for locomotion~\cite{kim2024not}.

\textbf{Partial observability.}
Real-world control is inherently a partially observable Markov decision process (POMDP). Compared to fully observable MDPs, learning under partial observability is more challenging, as policies must cope with state aliasing and incomplete information~\cite{kaelbling1998planning}. In practice, standard RL methods often exhibit degraded performance or unstable training when policies are restricted to partial observations~\cite{hausknecht2015deep,pinto2017asymmetric}. Several strategies have been proposed to mitigate these challenges, including large-scale parallelized RL training~\cite{rudin2022learning}, teacher–student distillation~\cite{lee2020learning}, and asymmetric actor–critic architectures where the critic has privileged state access~\cite{pinto2017asymmetric}.

\textbf{Reward design.}
Reward shaping is a central challenge, as locomotion policies often involve more than ten hand-crafted reward terms~\cite{lee2020learning,miki2022learning,ji2022concurrent,shin2023actuator}, making tuning difficult and requiring expert heuristics. Constraint-based formulations can reduce the dimensionality of reward terms~\cite{kim2024not}, but in practice, they shift part of the tuning complexity to the choice and scaling of constraints. Multi-morphology training similarly offloads tuning to central pattern generators~\cite{shafiee2024manyquadrupeds}. Other approaches attempt to automate reward design with large language models (Eureka~\cite{ma2023eureka}) or remove explicit rewards altogether by optimizing intrinsic objectives, such as DIAYN~\cite{eysenbach2018diversity}.

Most recent controllers~\cite{rudin2022learning,miki2022learning} achieve agility, relying on high-dimensional and ad hoc reward shaping. We instead formulate rewards directly from actuator energy losses, yielding a compact and physically meaningful objective. Our formulation only requires four reward terms, which reduces tuning complexity.

\subsection{Energy-efficient locomotion}

Energy-efficient control is critical for autonomous legged robots.
Total losses consist of controller-independent terms such as sensing, computation, and inverter switching, and motion- respectively control-policy-dependent losses, which can be optimized. The literature on modeling motion-dependent energy losses can be grouped into pseudo-approximations, low-fidelity models, and high-fidelity models. Here, \emph{low-fidelity} models capture only dominant, aggregated effects (e.g., mechanical power and Joule losses) using a small number of lumped parameters, often with hand-tuned scaling, enabling efficient policy optimization at the cost of limited physical detail.
In contrast, \emph{high-fidelity} models aim to represent individual loss mechanisms (e.g., copper, iron, magnet, and mechanical losses), either through detailed physics-based motor models or data-driven power predictors.
While such models can achieve higher accuracy, they typically incur substantially higher modeling or data requirements and increased computational cost, which can limit their direct use in large-scale reinforcement learning.

\textbf{Pseudo-approximations.}
Simplified proxies such as squared torque are frequently used to approximate energy costs. However, many works omit mechanical power consumption altogether~\cite{lee2020learning,shin2023actuator,miki2022learning,ji2022concurrent,kim2024not}, or introduce objectives with weak correlation to actual energy, e.g., penalizing changes in power~\cite{shafiee2024manyquadrupeds}.

\textbf{Low-fidelity models.}
These methods typically split energy into mechanical power~\cite{fu2021minimizing} and Joule heating~\cite{yang2022fast}. Scaling between these terms is often hand-tuned or derived from motor characteristics~\cite{wensing2017proprioceptive,fadini2021computational,mahankali2024maximizing}, enabling training of energy-aware locomotion controllers~\cite{aractingi2023controlling,roux2025constrained}.

\textbf{High-fidelity models.}
Advanced approaches numerically compute individual loss components, including copper losses ($P_\mathrm{Cu}$), iron losses ($P_\mathrm{FE}$), permanent magnet losses ($P_\mathrm{PM}$), and mechanical dissipation~\cite{ferrari2022flux}. Recent data-driven work has employed accurate power measurement to train neural networks that predict instantaneous power consumption in legged robots~\cite{valsecchi2024accurate}.

Existing energy models range from simplified proxies such as torque-squared penalties~\cite{lee2020learning,kim2024not} to computationally intensive numerical loss calculations~\cite{ferrari2022flux}. Building on prior actuator-aware formulations~\cite{wensing2017proprioceptive,fadini2021computational}, we propose a loss model tailored to \acp{pmsm} that captures the dominant sources of energy dissipation while remaining tractable within modern reinforcement learning simulators. Crucially, the resulting energy term is expressed in physical units [J], which avoids arbitrary reward scaling and allows the same coefficient to be reused across robot morphologies without retuning. This balance enables training of locomotion policies that are explicitly energy-efficient.

\subsection{Notation}
We adopt the following notation conventions throughout the paper:
\begin{itemize}
    \item Scalars are written normal $(x)$, vectors in bold $(\boldsymbol{x})$.
    \item Target values are denoted with a hat $(\hat{x})$.
    \item All vector norms are Euclidean, $\lVert \mathbf{x} \rVert = \lVert \mathbf{x} \rVert_2$.
    \item The term \emph{motor} refers to the actuator input side, and \emph{joint} refers to the output side.
    \item Reduced inertia with respect to the joint is indicated with a tilde $(\tilde{I})$.
    \item Robot legs are indexed LF (left front), RF (right front), LH (left hind), RH (right hind).
\end{itemize}

\section{Method}\label{sec:method}

Our pipeline to bridge the reality gap comprises three stages, illustrated in Figure~\ref{fig:approach_overview}. 
(i) \emph{Data collection on the real robot} (Section~\ref{sec:method:data_collection}), using joint impedance (PD) control to execute and record trajectories.
(ii) \emph{Simulator alignment via evolutionary parameter identification} (Section~\ref{sec:method:pace}), where we fit a compact set of parameters so that simulated in-air trajectories match the recorded ones. 
(iii) \emph{Policy learning and deployment} (Section~\ref{sec:method:control}), where we train a blind locomotion controller and deploy it zero-shot on hardware. 
We refer to the overall approach as \acf{pace}.

\subsection{Data collection}\label{sec:method:data_collection}

Data collection is conducted with a fixed base to avoid motion cross-coupling between different limbs.
We excite all joints simultaneously using chirp signals of \qtyrange{20}{60}{\second} per sequence, sweeping from a minimum to a maximum frequency. The chirp is applied at the \emph{joint position target} level and tracked by a PD controller, such that both the real system and the simulator are driven by identical reference trajectories rather than free-running dynamics. This eliminates accumulating phase error and makes direct trajectory matching well-posed. Signals and measurements are time-synchronized and logged at high sampling rates (typically \qtyrange{400}{10000}{\hertz}). We note that the conditions listed below were selected empirically for robustness and practicality; a systematic comparison of alternative data-collection strategies is beyond the scope of this work.

\paragraph{Fixed base (no base motion)}
During simulation replay we rigidly fix the base in air. 
For robots with symmetry planes (e.g., two for \robot{ANYmal}/\robot{Minimal}, one for \robot{Tytan}), we cancel net base wrenches by commanding symmetric joint trajectories. For robots without a symmetry plane, including fixed-base humanoids or robotic arms, the same identification procedure applies by rigidly constraining the base during excitation. The actuator-level identification itself remains unchanged.

\paragraph{No contacts (legs free in air).}
We avoid all contacts (including inter-leg) so the identification is not confounded by unmeasured external forces. Moreover, in stance the base inertia dominates the effective joint dynamics; collecting in-air data isolates leg/drive dynamics. A simplified model in Appendix~\ref{sec:appendix:effective_base_inertia} shows that in agile stance the effective base inertia exceeds leg/drive inertias by 1--2 orders of magnitude (\robot{ANYmal} example).

\paragraph{Excitation bandwidth.}
Ideally, trajectories cover up to $f_\text{policy}/2$ (Nyquist~\cite{tan2018digital} of the control policy), since this is the highest possible frequency that the controller can excite in the system. Structural constraints may limit this (e.g., \qty{2}{\hertz} on \robot{ANYmal}; \qty{10}{\hertz} on \robot{Tytan}/\robot{Minimal}). If structural or practical constraints prevent reaching this range, the excitation should at least cover twice the highest frequency expected in the locomotion controller’s motion (e.g., \qty{1}{\hertz} for our robots walking at \qty{1}{\meter\per\second}).

\paragraph{Joint-level PD gains.} \label{sec:method:data_collection:pd_gains}
We do not propose a formal or optimal procedure for selecting the PD
gains. In practice, gains are chosen conservatively based on experimental feasibility considerations.

The gains $P_\tau, D_\tau$ determine the closed-loop poles of the joint tracking dynamics, as given by the transfer function
\begin{align}
    H_q(s) \;=\; e^{-sT_d}\,\frac{P_\tau}{I_a s^2 + (d+D_\tau)s + P_\tau},
    \label{eq:method:transfer_function}
\end{align}
where $I_a$ denotes the effective armature inertia, $d$ the viscous damping, and $T_d$ a lumped delay.

Eq.~\eqref{eq:method:transfer_function} represents an \emph{idealized linear} closed-loop model and is used solely for frequency-domain analysis: it assumes sufficiently high-bandwidth inner-loop torque tracking, neglects saturation and other nonlinear effects, and is valid only when operating sufficiently far from actuator limits.

Increasing the gains shifts the dominant poles to higher frequencies, which would require substantially larger excitation bandwidths during data collection and are often infeasible on suspended hardware setups.

We therefore intentionally use comparatively low gains for both identification and policy training, such that the characteristic joint dynamics lie well within the achievable excitation range. This choice ensures that the dominant dynamics are observable and identifiable during data collection; empirically, higher gains did not improve identification quality but instead pushed the closed-loop poles beyond the feasible excitation bandwidth.

\subsection{Parameter identification} \label{sec:method:pace}

We align the simulator by fitting a small set of parameters that dominantly shape the joint-space dynamics: per-joint armature/inertia $\mathbf{I}_a$, viscous damping $\mathbf{d}$, Coulomb friction $\boldsymbol{\tau}_f$, and joint bias $\tilde{\mathbf{q}}_b$, plus a global command delay $T_d$. With $n$ actuated joints, the parameter vector is
\begin{align}
    \mathbf{p} \;=\; [\mathbf{I}_a,\;\mathbf{d},\;\boldsymbol{\tau}_f,\;\tilde{\mathbf{q}}_b,\;T_d]\tran \in \mathbb{R}^{4n+1}.
\end{align}
The joint bias $\tilde q_b$ captures quasi-static offsets between commanded and effective joint positions (e.g., encoder zeroing, assembly tolerances, or firmware compensations). These offsets are approximately constant over identification and deployment and manifest as steady-state PD set-point shifts if left unmodeled.

Figure~\ref{fig:background:blockdiagram_laplace} illustrates the full actuator and control stack for conceptual clarity, but it does not imply that all depicted subsystems are modeled or identified explicitly.
For the purpose of system identification, we deliberately operate at the \emph{joint-space, end-to-end} level and treat several inner control and electrical loops as ideal or absorbed by the parameterization (cf. Section~\ref{sec:discussion}).

Specifically, the inner current and torque control loops are assumed to exhibit sufficiently high bandwidth relative to the excitation range used for identification, such that they can be approximated as unity gain within this regime. Residual effects from firmware compensations, filtering, and unmodeled electronics are not parameterized individually but are instead captured implicitly through the effective inertia $I_a$, damping $d$, and lumped delay $T_d$.
This abstraction avoids introducing weakly identifiable parameters while preserving the dominant joint-space dynamics relevant for sim-to-real transfer.

\paragraph{On saturation effects.}
While actuator torque and velocity saturation are clearly present on the real systems, we do not include saturation limits as free parameters in the identification. Instead, saturation is treated as a \emph{known and enforced nonlinearity} that is applied deterministically in both simulation and hardware using manufacturer specifications and safety limits (cf. Figure~\ref{fig:torque_velocity_saturation_model} and Eq.~\eqref{eq:mass_damper_differential_equation}). As a result, the identified parameters capture the effective joint-space dynamics \emph{within} the feasible operating region, while saturation acts only as a boundary constraint. Including saturation limits as optimization variables would be poorly conditioned and weakly identifiable from in-air trajectories.

We instantiate $N=4096$ parallel environments, indexed by $e$, with the real-experiment base pose, each with parameters $\mathbf{p}_e$. We replay the recorded joint targets $\mathbf{\hat{q}}$ at the simulation rate used later for \ac{rl}, and measure the simulated joint trajectories $\mathbf{q}_{i,e}^{\text{sim}}$. The identification objective for environment $e$ is the time-averaged mean-squared joint-position error:
\begin{align}
    \ell_e \;=\; \frac{1}{k}\sum_{i=1}^{k}\|\mathbf{q}_{i}^{\text{real}} - \mathbf{q}_{i,e}^{\text{sim}}\|^2,
\end{align}
yielding a loss vector $\boldsymbol{\mathcal{L}} \in \mathbb{R}^{N}$. We optimize
\begin{align}
    \mathbf{p}^* \;=\; \argmin_{\mathbf{p}} \, \mathbb{E}[\ell_e],
\end{align}
\rev{using \ac{cmaes}~\cite{nomura2024cmaes} over the population. We chose \ac{cmaes} because the identification objective is trajectory-level, nonconvex, and affected by nonsmooth effects such as friction, saturation, and delay, while the optimizer does not require gradients through the full hardware-matched simulation pipeline. Compared with simpler evolutionary strategies, covariance adaptation provides an efficient search distribution for the moderately dimensional continuous parameter spaces considered here (typically $\approx 49$ parameters for our robots). We found \ac{cmaes} reliable and sample-efficient in massively parallel GPU simulation; for lower-dimensional or more expensive settings, Bayesian optimization is an alternative. We emphasize that \ac{cmaes} is used here as a generic black-box optimizer for parameter identification.
Closely related sim-to-real pipelines that also employ \ac{cmaes} or evolutionary strategies for simulator alignment have been reported previously~\cite{miller2025high,sobanbabu2025sampling}.}

\paragraph{Single-joint dynamics model.}
For reference, a single joint obeys
\begin{align}
    I_a \ddot{q} + d\,\dot{q} \;=\; \tau_i + \tau_\text{comp} + \tau_f, \label{eq:mass_damper_differential_equation}
\end{align}
where $\tau_f$ models Coulomb friction and $\tau_\text{comp}$ denotes firmware-level compensations (e.g., cogging compensations, plant inversion, friction observers). These terms are not modeled explicitly, identified, or used elsewhere in the pipeline; instead, their net effect is absorbed implicitly into the fitted parameters through trajectory matching.

Before introducing the closed-loop form in Eq.~\eqref{eq:method:eom_single_joint}, we recall that
$q$ denotes the joint position, $\hat q$ the commanded joint target,
$P_\tau,D_\tau$ the joint-level PD gains (Sec.~2.1), and $\tilde q_b$ a
constant joint-position bias.

Assuming tight current control and a saturation nonlinearity on commanded torque (cf. Figure~\ref{fig:torque_velocity_saturation_model}), a practical closed-loop form is
\begin{align}
    I_a \ddot{q} + d\,\dot{q}
    \;=\; \mathrm{sat}\!\big(P_\tau(\hat{q}-q+\tilde{q}_b) - D_\tau \dot{q} + \tau_\text{comp}\big) + \tau_f . \label{eq:method:eom_single_joint}
\end{align}
Drive-level simplifications (e.g., load-independent damping) are common but imperfect; actual damping and torque–current maps are often state dependent. Hence, fitting parameters against full-robot in-air data (rather than isolated drives; see Section~\ref{sec:results:full_robot}) is critical.

The parameter set $\mathbf{p}$ reflects the quantities that dominantly shape the closed-loop joint-space dynamics in Eq.~\eqref{eq:method:eom_single_joint} and Eq.~\eqref{eq:method:transfer_function} and are directly observable from encoder-only in-air trajectories under PD position control.
Specifically, the effective armature inertia $I_a$ and viscous damping
$d$ determine the dominant poles of the closed-loop system, Coulomb
friction $\tau_f$ captures low-velocity asymmetries, the joint bias
$\tilde q_b$ accounts for constant offsets due to encoder alignment and
mechanical preload, and the lumped delay $T_d$ captures communication and
inner-loop latencies.

Crucially, these parameters enter the dynamics in a manner that is
structurally identifiable from joint position measurements, whereas
many lower-level actuator parameters are not.

Finally, note a non-uniqueness if PD gains are co-optimized with the dynamics. Any common scaling $u_c$ of $\{I_a,d,P_\tau,D_\tau\}$ preserves Eq.~\eqref{eq:method:eom_single_joint} and trajectories:
\begin{align}
    u_c I_a^* \ddot{q} + u_c d^* \dot{q}
    \;=\; u_c P_\tau^* \Delta q - u_c D_\tau^* \dot{q}. \label{eq:method:non_uniqueness}
\end{align}
We therefore \emph{do not} include PD gains in the identification and assume these to be known.

The same structural non-uniqueness motivates excluding other multiplicative actuator-side parameters, such as the motor torque constant $k_i$, from the identification.
Under PD position control, $k_i$ only enters as a global torque scaling and is therefore not independently identifiable from joint position trajectories; its effect is absorbed into the effective inertia and damping terms.

\subsection{Learning environment}\label{sec:method:control}

Having addressed simulator alignment through parameter identification, we now turn to the learning-based control stage. Because \ac{pace} identifies the dynamics end-to-end, we \emph{do not} use dynamics randomization. We randomize the \emph{task} (pushes, ground friction) and terrains (flat/rough, stairs, boxes, slopes). We otherwise follow standard practice~\cite{rudin2022learning}, emphasizing only differences below.

\subsubsection{Observations}

We model the problem as a POMDP. Accordingly, we use asymmetric \ac{ppo}~\cite{pinto2017asymmetric}: the policy receives proprioception $\mathbf{o}^\text{prop}\subset\mathbf{s}$, while the critic observes privileged state $\mathbf{s}$.

\textbf{Policy (proprioception).} Base linear velocity $\mathbf{v}_B$, base angular velocity $\boldsymbol{\omega}_B$, gravity in base frame $\mathbf{g}_B$ (from a state estimator); user commands $(v_B^x,v_B^y,\omega_B^\text{yaw})$; joint positions $\mathbf{q}$ and velocities $\dot{\mathbf{q}}$; previous action $\mathbf{a}_{t-1}$. We add i.i.d.\ noise to all but commands and $\mathbf{a}_{t-1}$. The observation has dimension $\mathbf{o}^\text{prop}\in\mathbb{R}^{48}$.

\textbf{Critic (privileged).} Noise-free $\mathbf{o}^{\text{prop}*}$ plus base wrench $(\mathbf{F}_B,\boldsymbol{\tau}_B)$, ground friction, binary foot contacts, and a height scan centered at the base CoM spanning \SI{2}{\meter}$\times$\SI{3}{\meter} at \SI{0.15}{\meter} resolution. Thus $\mathbf{o}^\text{priv}\in\mathbb{R}^{305}$ and $\mathbf{s}=[\mathbf{o}^{\text{prop}*},\,\mathbf{o}^\text{priv}]\in\mathbb{R}^{353}$.

\subsubsection{Actions and hard-limit safe PD control}\label{sec:method:safe_pd_control}

The policy outputs joint position offsets $\mathbf{a}_t=\pi(\mathbf{o}^\text{prop}_t)$ relative to a default posture $\mathbf{q}_0$, which are converted to torques and tracked by a current loop (cf. Figure~\ref{fig:background:blockdiagram_laplace}):
\begin{align}
    \boldsymbol{\tau}_t \;=\; P_\tau \big(\mathbf{a}_t + \mathbf{q}_0 - \mathbf{q}\big) \;-\; D_\tau \dot{\mathbf{q}}.
\end{align}
Before sending targets $\hat{\mathbf{q}}_t=\mathbf{a}_t+\mathbf{q}_0$ to the drives we apply a \emph{joint-limit saturation} to guarantee zero commanded torque toward a hard limit while preserving motion away from it. 
Let $\mathcal{Q}_\text{feas}$ be the feasible set and $\mathcal{Q}_\text{soft}\subset\mathcal{Q}_\text{feas}$ a soft-limit band. For joint $j$ with soft/hard bounds $q_j^\text{soft},q_j^\text{hard}$,
\begin{align}
\hat{q}_j \;=\;
\begin{cases}
\hat{q}_j - \tfrac{q_j - q_j^\text{soft}}{q_j^\text{hard} - q_j^\text{soft}}\big(\hat{q}_j - q_j^\text{hard}\big),
& q_j\!\in\!\mathcal{Q}_\text{soft},~\hat{q}_j\!\notin\!\mathcal{Q}_\text{feas},\\[6pt]
\hat{q}_j, & \text{otherwise}.
\end{cases}
\end{align}
At $q_j=q_j^\text{hard}$ this yields $\hat{q}_j=q_j^\text{hard}$ so the PD term toward the limit vanishes. The same logic applies near the lower bound. This saturation runs in both simulation and on hardware.

\paragraph{Clarification of joint-limit saturation.}
The joint-limit handling does not constrain the policy action space nor prevent targets from exceeding the hard limits.
Instead, the commanded target $\hat{q}_j$ is reshaped such that the PD controller produces zero torque \emph{toward} the hard
limit while preserving full authority away from it.
Consequently, even with low gains, the controller can still generate large torques near the limits in the direction that
moves the joint back into the feasible set.

The soft-limit band $Q_\text{soft}$ is a narrow buffer (typically a few degrees, $\mathcal{O}(2$--$5^\circ)$ per joint)
introduced solely to smoothly interpolate the saturation and avoid discontinuities.
It does not represent an additional constraint or penalty and was found not to induce avoidance behaviors in practice.

\subsubsection{Rewards}\label{sec:method:rewards}

We define the number of reward terms by the number of \emph{independently tuned scalar weights} that must be manually chosen.
Accordingly, although some objectives may decompose into multiple physical components, they are counted as a single reward term if no additional relative scaling is introduced.
Thanks to fitted dynamics parameters, we therefore use only four reward terms: velocity tracking and energy (physics-based), plus two structural penalties (collisions and foot-touchdown velocity).
In particular, the energy objective combines electrical, mechanical, and gravitational contributions through a physics-grounded formulation with a single weight, avoiding the need for hand-tuning relative scalings that are commonly treated as separate rewards in prior work (e.g. vertical velocity penalty, torque-squared).

\paragraph{Velocity tracking.}
Following~\cite{lee2020learning}, we reward proximity to commanded base velocities:
\begin{align}
    \boxed{
    r_v \;=\; 
    \exp\!\Big(-\tfrac{\|\hat{\mathbf{v}}_{B,xy}-\mathbf{v}_{B,xy}\|_2^2}{\sigma_v}\Big)
    + \exp\!\Big(-\tfrac{(\hat{\omega}_{B,\text{yaw}}-\omega_{B,\text{yaw}})^2}{\sigma_v}\Big)
    }.
\end{align}

\paragraph{Energy.}
We combine electrical dissipation $P_\text{el}$ with mechanical power $P_\text{mech}$ (including regeneration) and gravitational potential power $P_\text{pot}$:
\begin{align}
    P_\text{total} \;=\; P_\text{el} + P_\text{mech} + P_\text{pot}.
\end{align}
Assuming dominant $q$-axis current $i_q$ and negligible $i_d$,
\begin{align}
    P_\text{el} &= \sum_{j=1}^n R_j i_{q,j}^2 
               \;=\; \sum_{j=1}^n \tau_j^2\,\frac{R_j}{r_j^2 k_{i,j}^2}, \label{eq:joule_heating} \\
    P_\text{mech} &= 
    \begin{cases}
    \boldsymbol{\tau}^\top \dot{\mathbf{q}}, & \boldsymbol{\tau}^\top \dot{\mathbf{q}} > 0,\\
    k_\text{regen}\,\boldsymbol{\tau}^\top \dot{\mathbf{q}}, & \boldsymbol{\tau}^\top \dot{\mathbf{q}} < 0,
    \end{cases}\\
    P_\text{pot} &= \sum_{b=1}^{B} m_b g\, v_{b,z},
\end{align}
where $r_j$ is the gear ratio, $k_{i,j}$ the motor constant, $R_j$ the coil resistance, $g$ gravity, and $v_{b,z}$ the center-of-mass velocity along $-\,\mathbf{g}$. Because many robotic systems are black-box, Figure~\ref{fig:appendix:joule_heating_constant} reports the characteristic Joule heating scale from Eq.~\eqref{eq:joule_heating} for our in-house developed robots, providing a rough estimate across different platforms.

Since joint speeds scale with base speed, dissipation grows roughly with $\|\hat{\mathbf{v}}_B\|^2$. Without normalization, the physically derived power term would therefore dominate the reward at higher commanded velocities and bias the policy toward under-tracking. We thus apply a velocity-dependent normalization,
\begin{align}
    \gamma_v \;=\; \frac{1}{\|\hat{\mathbf{v}}_B\|^2 + 1},
\end{align}
and define
\begin{align}
    \boxed{r_e \;=\; \gamma_v\, P_\text{total}},
\end{align}
which ensures balancing between velocity tracking and energy optimization across different commanded speeds while preserving the physical interpretation of the power-based loss. Alternative formulations based on constrained or complementary optimization could enforce velocity tracking and energy minimization separately; however, the proposed normalization provides an equivalent effect in a simpler unconstrained reward formulation without introducing additional constraints or dual variables.

\paragraph{Interpretation of energy terms.}
While some components of the energy objective resemble commonly used heuristic penalties, their interpretation and effect differ fundamentally.
In particular, the gravitational potential term $P_{\mathrm{pot}}$ is formulated as a \emph{reward}, not a penalty.
It rewards increases in potential energy when ascending (e.g., stairs or slopes) and, by energy conservation, discourages unnecessary energy injection when descending.
This behavior cannot be reproduced by a simple vertical velocity penalty, which lacks directionality and physical grounding.

Crucially, all energetic terms -- electrical dissipation, mechanical power, and gravitational potential -- share the same physical unit [J].
The resulting reward therefore operates in a unified energy space, where scaling is determined by physical constants (e.g., mass, gravity, motor parameters) rather than ad-hoc coefficients.
This ensures consistent weighting across robots and operating regimes, and shifts the design effort from manual tuning toward physically meaningful normalization.

\paragraph{Limitations of the energy model.}
\rev{
The proposed energy model captures the dominant motion-dependent losses relevant for large-scale policy optimization, but it is not a full motor-drive loss model. In particular, it does not explicitly model iron losses, inverter switching losses, temperature-dependent resistance, battery dynamics, detailed field-weakening behavior, or load-dependent transmission losses. The model is therefore expected to be most accurate for \ac{pmsm}-driven robots operating within the identified torque, velocity, and temperature regimes, and less accurate for systems with strongly nonlinear transmissions, pronounced thermal drift, or frequent operation close to electrical saturation. In such cases, the same framework could incorporate higher-fidelity or data-driven power models, at the cost of additional modeling and identification effort.}

\paragraph{Foot-touchdown (FTD) penalty.}
To discourage braking by impacts (gear wear, noise), we penalize the maximum foot speed in a short history window upon touchdown. With buffer length $n_\text{ftd}=3$ and foot $j$,
\begin{align}
    v_{j,\text{ftd}} &=
    \begin{cases}
    \max_{k\in\{t-2,t-1,t\}} \|\mathbf{v}^{\text{foot}}_{j,k}\|, & \text{if touchdown at } t,\\
    0, & \text{otherwise},
    \end{cases}\\
    \Aboxed{r_\text{ftd} &= \sum_{j}^{\#\text{feet}} v_{j,\text{ftd}} }.
\end{align}

While penalizing foot jerk is a reasonable alternative and can promote smooth motion, it implicitly acts at a fixed temporal scale determined by the control timestep and numerical differentiation. In this work, our goal was to explicitly target impact-related effects at contact, for which a windowed foot-touchdown velocity penalty offers more direct temporal control and interpretability. We therefore opted for the FTD formulation.

\paragraph{Collision penalty.}
We penalize joint-limit and thigh–environment collisions via an indicator:
\begin{align}
    \boxed{
    r_c \;=\;
    \begin{cases}
    1, & \text{if collision at } t,\\
    0, & \text{otherwise}.
    \end{cases}}
\end{align}

\paragraph{Penalty scheduling.}
Early in training, large penalties can stall gait discovery. We therefore schedule energy and FTD penalties with an exponential factor $k_\text{decay}=e^{-\lambda t}$ and $\kappa=1-k_\text{decay}$:
\begin{align}
    \Aboxed{ r \;=\; c_v r_v + c_c r_c + \kappa\,(c_e r_e + c_\text{ftd} r_\text{ftd}) }.
\end{align}
We typically choose a half-life of 500 iterations for $\lambda$, but this is task dependent.

\subsubsection{Entropy scheduling}

Exploration is essential early but harms precision later. We anneal the entropy coefficient with a smooth tanh schedule:
\begin{align}
    \mathcal{E}(t) &= \mathcal{E}_\infty + \epsilon\,(\mathcal{E}_0 - \mathcal{E}_\infty),\\
    \epsilon &= \tfrac{1}{2} - \tfrac{1}{2}\tanh\!\big(\eta\,(t - T_\mathcal{E})\big),
\end{align}
so that $\mathcal{E}(0)\!\approx\!\mathcal{E}_0$ transitions to $\mathcal{E}_\infty$ near $T_\mathcal{E}$.

\subsection{Remarks} \label{sec:method:remarks}

\textbf{Drive-side filtering.} Joint velocities are often low-pass filtered (cutoffs as low as \SI{25}{}–\SI{50}{\hertz}). Although \ac{pace} uses positions in the loss, filtered $\dot{q}$ still affects commanded PD torques. If noticeable, include this filter in simulation and identify its parameters.

\textbf{Units of PD gains.} Non-\si{SI} units still allow identification but break generalization across gains and skew energy terms (damping/ohmic vs.\ mechanical). Use consistent \si{SI} units.

\begin{figure}
    \centering
    \includegraphics[width=\linewidth]{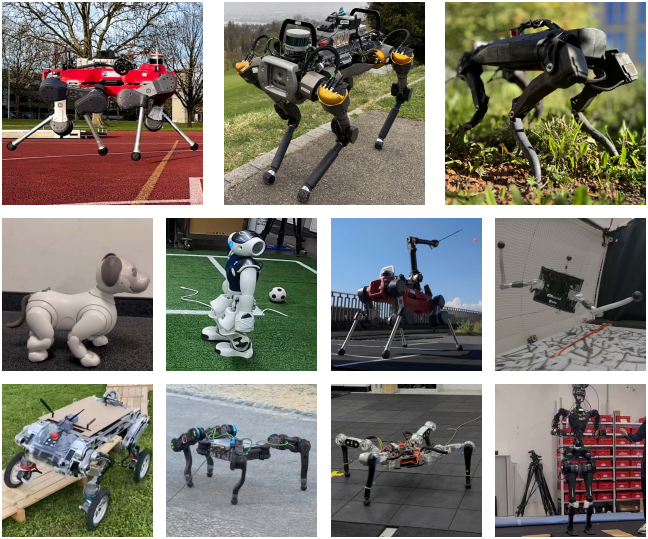}
    \caption{Primary robotic platforms evaluated in this study. Top row (left to right): \robot{ANYmal}, \robot{Tytan}, and \robot{Minimal}. Additional systems include Sony's \robot{Aibo}, Softbank's \robot{NAO}, \robot{ALMA}, \robot{Spacehopper}, \robot{LEVA}, \robot{Magnecko} v2 and v1, and Fourier's \robot{GR-1}. All robots are depicted in operation using the proposed \ac{pace} parameter fitting.}
    \label{fig:experiments:robots_using_pace}
\end{figure}

\textbf{Further assumptions (empirically satisfied in Figure~\ref{fig:experiments:robots_using_pace}).}
\begin{itemize}
    \item Correct kinematics (URDF/USD, frames).
    \item High-bandwidth current control or an LTI-approximable drive (closed-loop behavior is fitted).
    \item Mild temperature dependence during data collection.
    \item Nonlinearities modest in the excited range (or absorbed by fitted terms).
    \item Sufficient structural stiffness in the excitation band (Section~\ref{sec:method:data_collection}).
\end{itemize}

Finally, we stress that adding parameters indiscriminately can harm identifiability. The compact set $\{\mathbf{I}_a,\mathbf{d},\boldsymbol{\tau}_f,\tilde{\mathbf{q}}_b,T_d\}$ proved sufficient across platforms; co-optimizing PD gains leads to non-unique optima (Eq.~\eqref{eq:method:non_uniqueness}).

\section{Experiments}\label{sec:experiments}
This section introduces the robotic platforms and the experimental protocols. We evaluate \ac{pace} in two stages.
First, controlled in-air validation and evaluation experiments (Section~\ref{sec:experiments:model_analysis}) establish methodological soundness at different levels of system complexity, ranging from a single drive to the fully suspended robot. 
Second, on-ground locomotion experiments (Section~\ref{sec:experiments:sim2real}) apply the approach in realistic deployment scenarios, assessing performance and energetic efficiency.

We proceed bottom–up: 
(In–air: i) single–actuator analysis on \robot{Tytan} (full access to firmware, electronics, and mechanics, Section~\ref{sec:method:single_drive}); 
(In–air: ii) full–robot identification and validation, including a comparison on \robot{ANYmal} against a zero–model baseline and a state-of-the-art actuator network~\cite{hwangbo2019learning} (Section~\ref{sec:experiments:full_robot}); 
(On–ground: iii) full–robot locomotion experiments (Sections~\ref{sec:experiments:sim2real:tytan}, \ref{sec:experiments:sim2real:anymal}) under the same parameters as (In–air: ii) and
(On–ground: iv) long-duration energetic evaluations (Section~\ref{sec:experiments:running_track}).

\paragraph{Separation of in-air identification and on-ground evaluation.}
For clarity, we emphasize that all parameter identification in this work is performed exclusively using \emph{in-air, contact-free} trajectories with a fixed or effectively immobilized base (Section~\ref{sec:experiments:model_analysis}). No data collected under ground contact is used for parameter estimation. Ground-contact experiments (Section~\ref{sec:experiments:sim2real}) are used solely for downstream policy deployment and evaluation using the already identified parameters from Section~\ref{sec:experiments:full_robot}.

Each experimental stage has a distinct objective. 
The single–drive study validates high-bandwidth motor torque tracking and mechanical identification under fully known conditions. 
The \emph{in-air} full–robot experiments test whether the approach scales to system level, generalizes across PD gains and trajectories, and allow benchmarking against a zero–model baseline and a learned actuator network~\cite{hwangbo2019learning}.
Finally, the \emph{on–ground} locomotion trials demonstrate the practical applicability of \ac{pace}, evaluating tracking accuracy, energetic efficiency, and long–duration performance.

\begin{table}
\small\sf\centering
\caption{Main robot characteristics.}
\label{tab:experiments:robot_characteristics}
\begin{tabular}{c|c|c|c|c}
\toprule
& \multicolumn{2}{c|}{\robot{Tytan}} & \robot{ANYmal} & \robot{Minimal} \\
& Hips & Knee & & \\
\hline\hline
Weight [kg] & \multicolumn{2}{c|}{\num{52.3}} & \num{52.8} & \num{4.2} \\
Shoulder height [m] & \multicolumn{2}{c|}{\num{0.62}} & \num{0.55} & \num{0.25} \\
Shoulder width [m] & \multicolumn{2}{c|}{\num{0.22}} & \num{0.20} & \num{0.16} \\
Shoulder depth [m] & \multicolumn{2}{c|}{\num{0.73}} & \num{0.75} & \num{0.39} \\
Thigh length [m] & \multicolumn{2}{c|}{\num{0.40}} & \num{0.30} & \num{0.15} \\
Shank length [m] & \multicolumn{2}{c|}{\num{0.37}} & \num{0.38} & \num{0.16} \\
Regen.\ coeff.\ $k_\text{regen}$ [-] & \multicolumn{2}{c|}{\num{0.3}} & \num{0.0} & \num{0.3} \\
Bus voltage $u$ [V] & \multicolumn{2}{c|}{\num{48}} & \num{48} & \num{18} \\
\multirow{2}{*}{Gear ratio $r$ [-]} & \multirow{2}{*}{\num{5.6}} & \num{0.8} - & \multirow{2}{*}{?} & \num{7.2} - \\
& & \num{9} & & \num{16} \\
\multirow{2}{*}{Max.\ joint torque [Nm]} & \multirow{2}{*}{\num{140}} & \num{28} - & \multirow{2}{*}{\num{89}} & \num{2.9} - \\
& & \num{315} & & \num{6.4} \\
\multirow{2}{*}{Max.\ joint speed [rad/s]} & \multirow{2}{*}{\num{16.8}} & \num{3} - & \multirow{2}{*}{\num{8.5}} & \num{45} - \\
& & \num{36} & & \num{99.4} \\
Max.\ motor torque [Nm] & \num{25} & \num{35} & ? & \num{0.4} \\
Max.\ motor speed [rad/s] & \num{94} & \num{29} & ? & \num{716} \\
Motor constant $k_i$ [Nm/A] & \num{0.59} & \num{1.25} & ? & \num{0.0252} \\
Coil resistance $R$ [$\Omega$] & \num{1.04} & \num{1.71} & ? & \num{0.194} \\
\bottomrule
\end{tabular}
\end{table}

\begin{table}
\small\sf\centering
\caption{Reward scales and PD gains used for locomotion.}
\label{tab:experiments:reward_scales}
\begin{tabular}{c|c|c|c}
\toprule
& \robot{Tytan} & \robot{ANYmal} & \robot{Minimal} \\
\hline\hline
Velocity tracking & \num{0.2} & \num{0.2} & \num{0.2} \\
Energy [\num{e-5}] & $-\,\num{16}$ & $-\,\num{16}$ & $-\,\num{128}$ \\
Collisions & $-\,1.0$ & $-\,1.0$ & $-\,1.0$ \\
FTD & $-\,0.1$ & $-\,0.1$ & $-\,0.1$ \\
\hline
$P_\tau$ & $\,60$ & $\,85$ & $\,4$ \\
$D_\tau$ & $\,2$ & $\,0.6$ & $\,0.05$ \\
\bottomrule
\end{tabular}
\end{table}

\subsection{Robots}

We perform analysis on three quadrupeds--\robot{Tytan}, \robot{ANYmal}, and \robot{Minimal}--chosen to span actuation types, scales, and transparency (open vs. closed source). 
All three share the same leg topology: Hip Abduction–Adduction (HAA), Hip Flexion–Extension (HFE), and Knee Flexion–Extension (KFE), with point feet (rubber end caps).
All joints are \ac{pmsm} driven. 
Key characteristics are summarized in Table~\ref{tab:experiments:robot_characteristics}, where
``?'' marks indicate confidential specifications we are not permitted to disclose.

\noindent\textbf{\robot{Tytan}}—a custom platform developed at ETH Zurich and based on \robot{Barry}~\cite{valsecchi2023barry}—uses pseudo-direct drives at HAA/HFE (fixed ratio $r\!=\!5.6$) and a variable-ratio ball-screw lever at KFE (\(r \in [0.8, 9]\)). It employs a low inertia leg design by moving the knee-motor towards its hip.

\noindent\textbf{\robot{ANYmal}}~\cite{hutter2016anymal} employs series-elastic harmonic drives with high, fixed ratios. It serves as a closed-source testbed demonstrating applicability when only limited low-level access is available. Our method does \emph{not} rely on joint-torque sensors, in contrast to the actuator network baseline.

\noindent\textbf{\robot{Minimal}} is a small, largely 3D-printed quadruped. All joints share a variable-ratio lever mechanism driven by T-Motor units and a lead-screw transmission. In total \qty{76}{\percent} of its total mass is situated in the base with only \qty{6}{\percent} at each leg.

Beyond these three primary platforms, \ac{pace} has also been deployed on a diverse set of robots, including \robot{Aibo}~\cite{watanabe2025learning}, \robot{NAO}, \robot{ALMA}~\cite{bellicoso2019alma, ma2025learning}, \robot{Spacehopper}~\cite{spiridonov2024spacehopper}, \robot{LEVA}~\cite{arnold2025leva}, \robot{Magnecko}~v1~\cite{leuthard2024magnecko}/v2, \robot{GR-1}~\cite{he2025attention}, the \robot{Allegro} Hand, and the \robot{Ability} Hand (see Fig.~\ref{fig:experiments:robots_using_pace}), as well as additional unpublished systems not shown here.

\begin{figure*}
    \centering
    \begin{subfigure}{0.3\linewidth}
        \centering
        \includegraphics[height=2.401in]{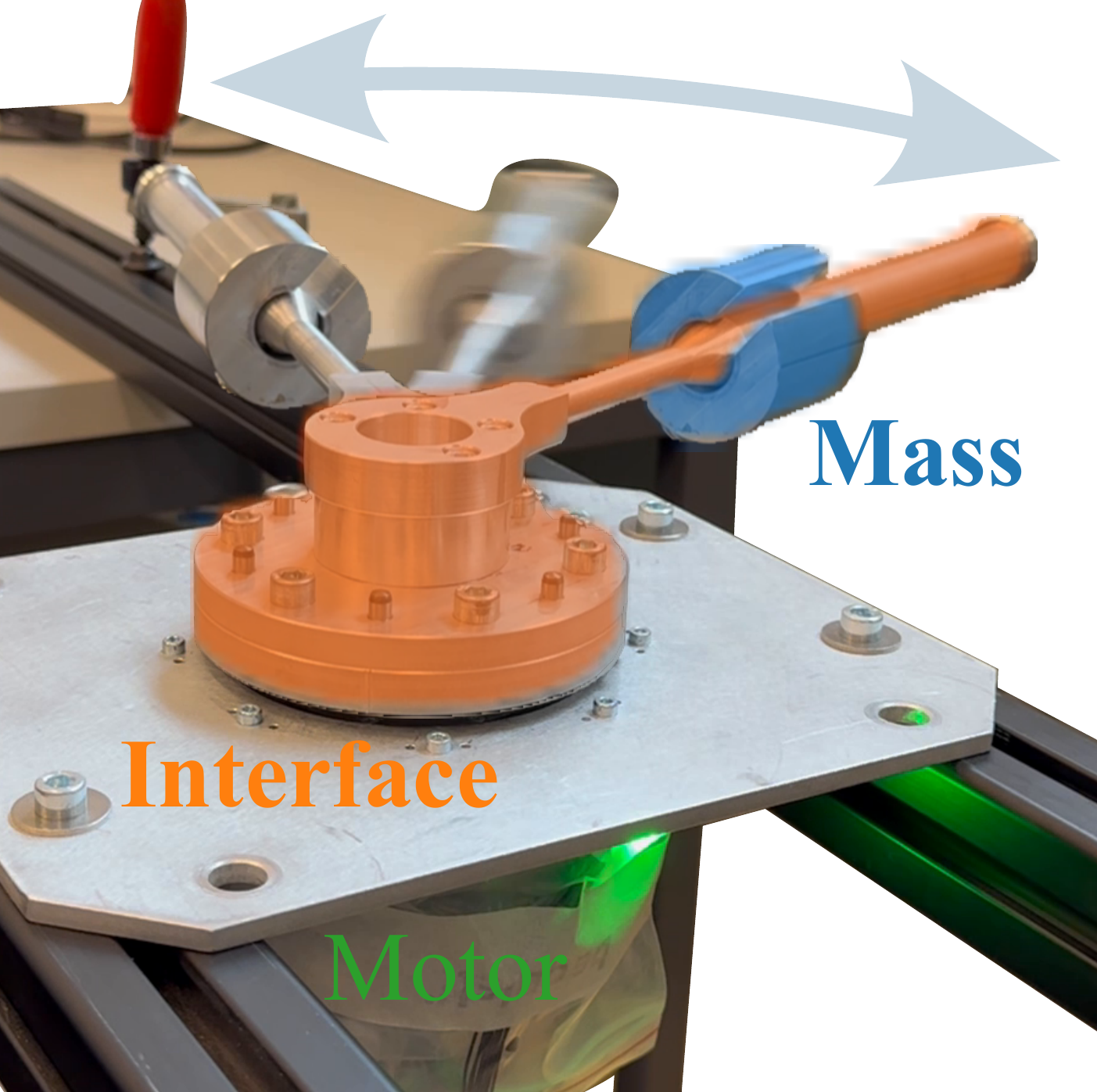}
        \caption{Single-drive experimental setup for joint-level characterization.}
        \label{fig:experiments:single_drive_stills}
    \end{subfigure}\hfill
    \begin{subfigure}{0.6109\linewidth}
        \centering
        \includegraphics[width=\linewidth]{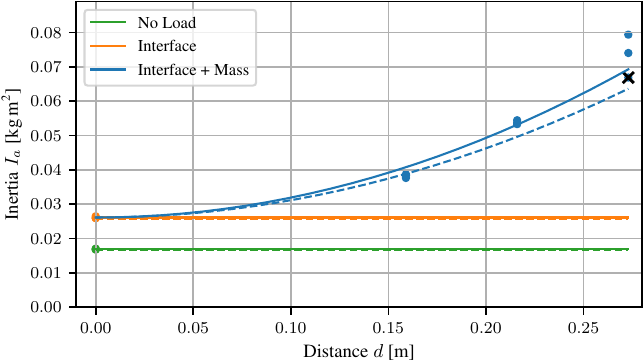}
        \caption{Identified versus target inertia as a function of lever arm radius, with drive compensations disabled.}
        \label{fig:results:single_drive}
    \end{subfigure}
    
    \vfill
    
    \begin{subfigure}{0.3\linewidth}
        \centering
        \includegraphics[height=2.401in]{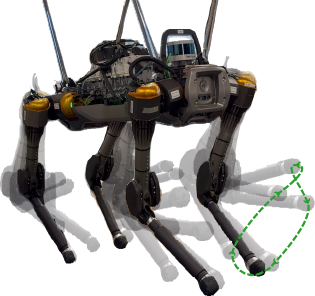}
        \caption{Full-robot setup on \robot{Tytan} for multi-joint data collection.}
        \label{fig:experiments:tytan_shake_stills}
    \end{subfigure}\hfill
    \begin{subfigure}{0.6109\linewidth}
        \centering
        \includegraphics[width=\linewidth]{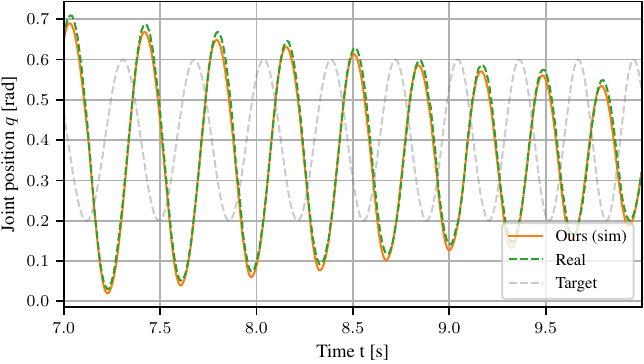}
        \caption{Trajectory replay on the LF HFE joint of \robot{Tytan}, comparing real and simulated joint positions.}
        \label{fig:results:tytan_hfe_trajectory_evaluation}
    \end{subfigure}
    
    \caption{Experimental setups and representative evaluation across abstraction levels.}
    \label{fig:experiments:single_drive_leg_robot}
\end{figure*}

\subsubsection{Data collection}

On each system, we record multi-joint chirps with varying amplitudes. Typical sequence duration is \qtyrange{20}{40}{\second} with $f_0=\qty{0.1}{\hertz}$ and a maximum frequency of \qty{10}{\hertz} to avoid excessive actuator stress. Structural constraints limit the achievable bandwidth on suspended setups (full \robot{Tytan}: \qty{8}{\hertz}; \robot{ANYmal}: \qty{2}{\hertz}; cf.\ Section~\ref{sec:method:data_collection}).

All full-robot logs (and low-level control) run at \qty{400}{\hertz}. We capture time-synchronized data via shared-memory logging using SignalLogger~\cite{anybotics_signal_logger}. 
For additional validation on \robot{Minimal}, we also collect random joint steps at \qty{2}{\hertz} update (every \qty{0.5}{\second}); we \emph{do not} use this trajectory on larger machines to avoid transmission wear.

Unless noted otherwise, all \ac{pace} optimizations and simulations run on a single NVIDIA GeForce RTX 3080.

\subsubsection{Initialization of CMA-ES}
All optimization parameters are normalized to lie within the interval $[-1,1]$ using their respective lower and upper bounds.
CMA-ES is initialized with a zero mean, corresponding to the center of the feasible parameter range and thus a non-biased prior over the parameters.
The initial standard deviation of the covariance matrix is set to $\sigma = 0.5$, enabling broad exploration of the normalized parameter space while remaining within the specified bounds.
This scale-invariant initialization is shared across all robots and parameter sets and was found to be robust in practice without requiring manual tuning.

\subsection{In-air evaluation and validation}\label{sec:experiments:model_analysis}

The following experiments validate \ac{pace} under controlled, contact-free conditions. 
The \textbf{single-drive setup} isolates mechanical effects and directly reveals the physical meaning of the identified inertia $I_a$ and damping $d$. We note that this setup is used purely for validation and is not required by the proposed identification pipeline, which only assumes the ability to command joint-level trajectories and log joint states through the standard robot interface.
The \textbf{in-air full-robot experiments} then test whether the approach scales when all actuators, electronics, and dynamics interact simultaneously. 
Throughout, we concentrate on effective inertia $I_a$ (mechanically verifiable); damping and friction are identified but not dissected further, as detailed tribology would require dedicated equipment.

\begin{figure*}
    \centering

    \begin{subfigure}{\linewidth}
        \centering
        \includegraphics[width=\linewidth]{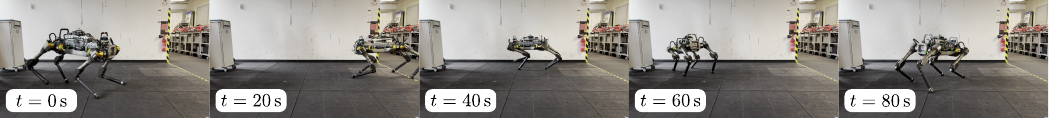}
        \caption{Real-world deployment of \robot{Tytan} during continuous deployment.}
        \label{fig:results:sim2real:tytan:real_stills}
    \end{subfigure}

    \begin{subfigure}{\linewidth}
        \centering
        \includegraphics[width=\linewidth]{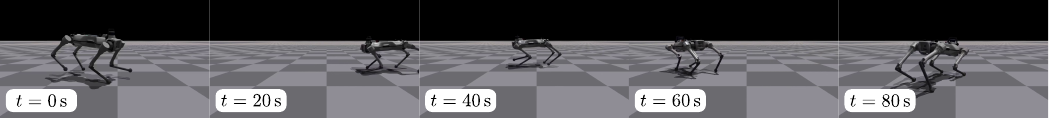}
        \caption{Corresponding simulation rollout under identical base velocity targets and conditions.}
        \label{fig:results:sim2real:tytan:sim_stills}
    \end{subfigure}

    \begin{subfigure}{\linewidth}
        \centering
        \includegraphics[width=\linewidth]{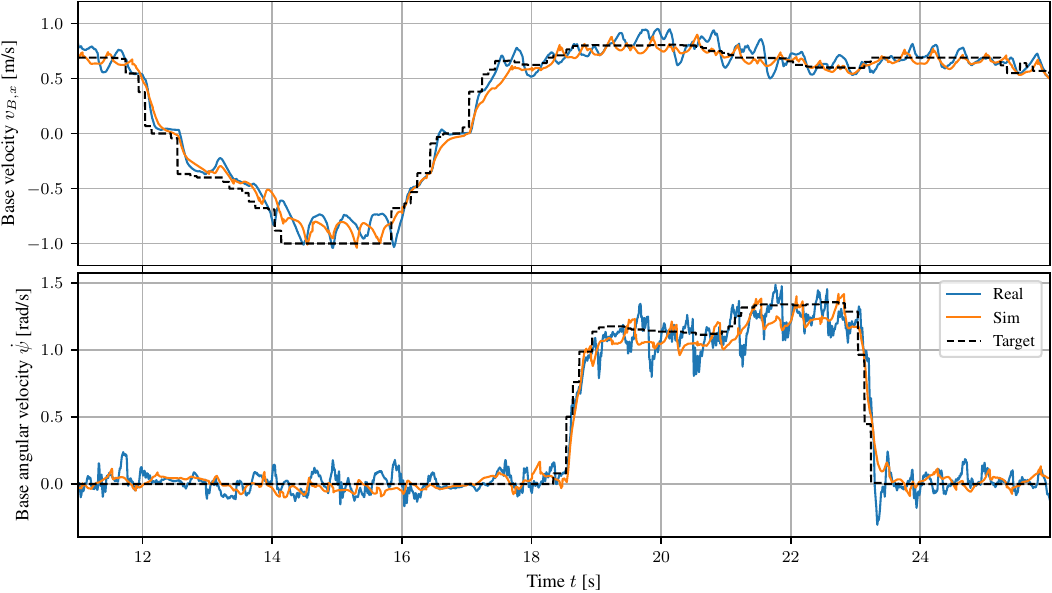}
        \caption{Tracking performance. Top: commanded and measured forward velocity. Bottom: commanded and measured angular velocity around the yaw axis.}
        \label{fig:results:sim2real:tytan:linear_angular_vel}
    \end{subfigure}

    \begin{subfigure}{\linewidth}
        \centering
        \includegraphics[width=\linewidth]{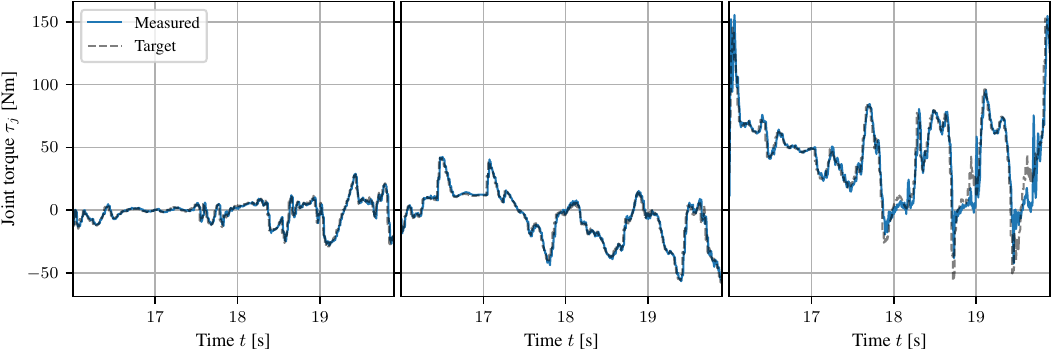}
        \caption{Joint torque tracking on the left front leg. Shown are policy-commanded and measured torques for the HAA, HFE, and KFE joints (left to right).}
        \label{fig:results:sim2real:tytan:torque_haa}
    \end{subfigure}

    \caption{Sim-to-real evaluation of \robot{Tytan} using the proposed \ac{pace} framework.}
    \label{fig:results:sim2real:tytan}
\end{figure*}

\subsubsection{Single drive}\label{sec:method:single_drive}

A single hip actuator is rigidly mounted to an aluminum frame (Figure~\ref{fig:experiments:single_drive_stills}). The motor shaft is vertical (gravity effects negligible). An aluminum interface attaches a discrete mass at known radii using a locking pin, enabling controlled changes of output inertia. The power supply is four \qty{12}{\volt} car-batteries in series (\qty{48}{\volt}), peak \qty{10.8}{\kilo\watt}; a \qty{100}{\ampere} fuse limits to \qty{4.8}{\kilo\watt}. The drive accepts up to \qty{32}{\ampere} peak current. This setup emulates a perfect voltage source.

The drive’s output-reduced inertia comprises the rotor, sun-gear assembly, planet gears, and output shaft. Using d’Alembert’s principle, the combined reduced output inertia is
\begin{align}
    \tilde{I}_\text{output}
    &= \tilde{I}_\text{rotor} + \tilde{I}_\text{sun} + \tilde{I}_\text{planet} + \tilde{I}_\text{shaft} \nonumber\\
    &=(1.14 + 0.511 + 0.00203 + 0.00146)\times 10^{-2}\; \si{\kilogram\metre\squared} \nonumber\\
    &\approx \SI{1.65e-2}{\kilogram\metre\squared},
    \label{eq:experiments:rotor_inertia}
\end{align}
showing rotor and sun-assembly dominance; planets and output-shaft are $\sim\!10^3$ smaller.

The interface inertia is $\tilde{I}_\text{interface}=\qty{8.67e-3}{\kilogram\metre\squared}$. The attachable mass has $m=\qty{503}{\gram}$ and own inertia about its CoM of \qty{1.8e-4}{\kilogram\metre\squared}. By varying the mounting radius of this mass, the joint inertia can be adjusted in the range $\qtyrange{1.65}{6.27e-2}{\kilogram\metre\squared}$.

\paragraph{Current loop validation.}
The goal of this experiment is to verify the motor drives’ ability to precisely track commanded currents $i_q$, and thereby commanded motor torques $\tau_m$, over a wide frequency range.
With the interface removed (free spin), we command a current chirp (\qty{1}{}–\qty{1250}{\hertz}, amplitude \qty{2}{\ampere}, \qty{25}{\second}) and record measured currents at \qty{10}{\kilo\hertz}. This experiment does not serve parameter identification, but rather confirms high-bandwidth motor torque authority. Joint-level torque fidelity is addressed end-to-end by \ac{pace}.

\begin{figure}
    \centering
    \includegraphics[width=\linewidth]{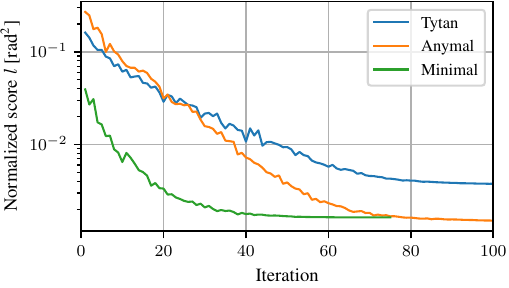}
    \caption{CMA-ES optimization score as a function of iteration (logarithmic scale) for the three main robotic platforms.}
    \label{fig:results:fitting_scores}
\end{figure}

\begin{figure}
    \centering
    \includegraphics[width=\linewidth]{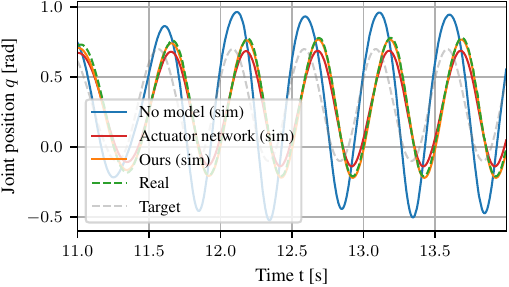}
    \caption{In-air LF–HFE joint trajectory of \robot{ANYmal} (zoomed view). Dashed gray: target trajectory. Dashed green: measured trajectory. Blue: no actuator model. Red: actuator network baseline. Orange: proposed method (\ac{pace}).}
    \label{fig:results:anymal_air_hfe_trajectory}
\end{figure}

\paragraph{Mechanical loop identification.}
For single-drive identification, we fit three parameters: armature inertia $I_a$, viscous damping $d$, and friction $\tau_f$. Joint bias is irrelevant (no gravity coupling), and the control/communication delay $T_d$ is negligible (microseconds). We simulate in Isaac Gym a single-link, single-joint model (link inertia set to zero so $I_a$ absorbs total inertia).

We collect chirps from \qtyrange{0.1}{10}{\hertz} at \qty{2.5}{\kilo\hertz} logging. 
In total $30$ experiments are conducted: $15$ with nominal firmware feed-forward compensations enabled (cogging/friction; cf.\ Figure~\ref{fig:background:blockdiagram_laplace}) and $15$ with compensations disabled. Each set spans five load cases (No load; Interface only; Interface+mass at \qty{15.9}{\centi\metre}, \qty{21.6}{\centi\metre}, \qty{27.3}{\centi\metre}) and three PD configurations:
\begin{itemize}\setlength\itemsep{0pt}
    \item $P_\tau=\qty{60}{\newton\metre\per\radian}$, $D_\tau=\qty{2}{\newton\metre\second\per\radian}$ (locomotion default),
    \item $P_\tau=\qty{145}{\newton\metre\per\radian}$, $D_\tau=\qty{5}{\newton\metre\second\per\radian}$,
    \item $P_\tau=\qty{250}{\newton\metre\per\radian}$, $D_\tau=\qty{10}{\newton\metre\second\per\radian}$.
\end{itemize}
Each experiment is fitted independently (three parameters), totaling $\sim\!\qty{5}{\hour}$ across all runs.

\begin{figure*}
    \centering

    \begin{minipage}{0.49\textwidth}
        \begin{subfigure}{\linewidth}
            \centering
            \includegraphics[width=\linewidth]{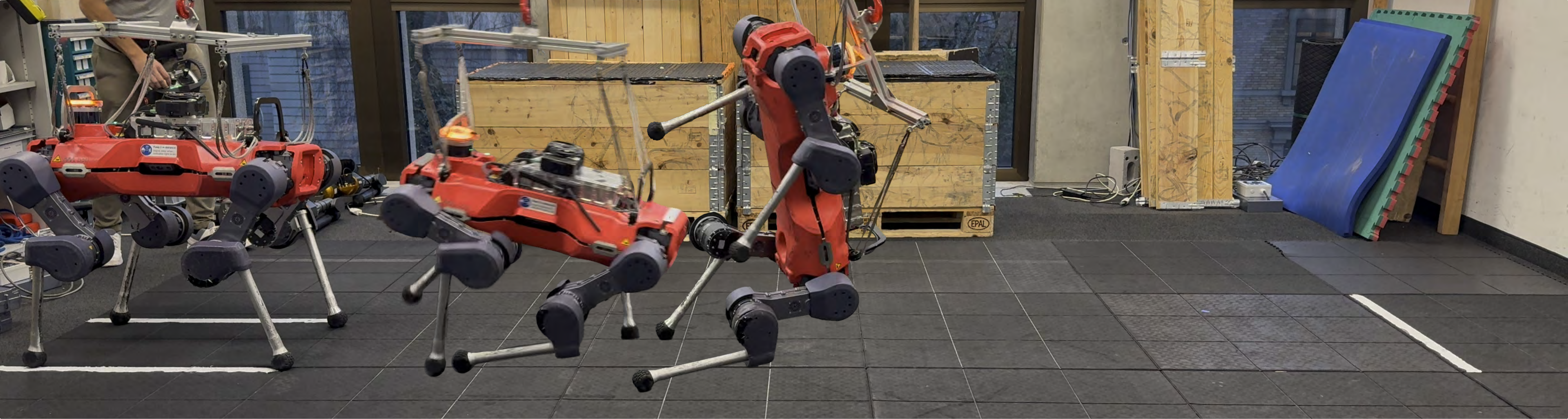}
            \caption{Real-world deployment without any actuator model.}
            \label{fig:results:sim2real:anymal:stills_nothing}
        \end{subfigure}

        \vspace{0.6em}

        \begin{subfigure}{\linewidth}
            \centering
            \includegraphics[width=\linewidth]{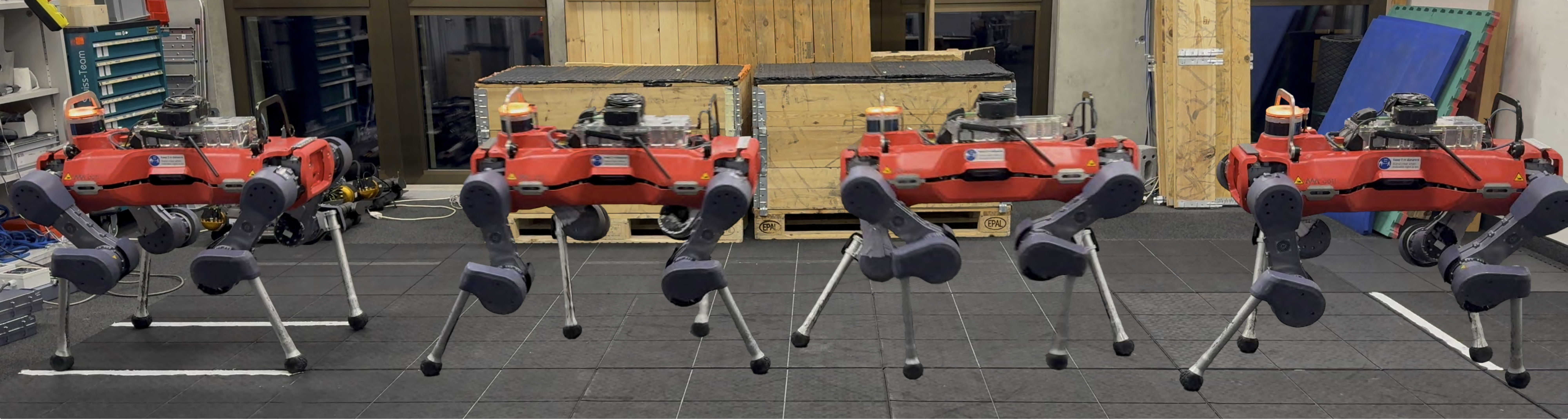}
            \caption{Real-world deployment with a learned actuator network.}
            \label{fig:results:sim2real:anymal:stills_act_net}
        \end{subfigure}

        \vspace{0.6em}

        \begin{subfigure}{\linewidth}
            \centering
            \includegraphics[width=\linewidth]{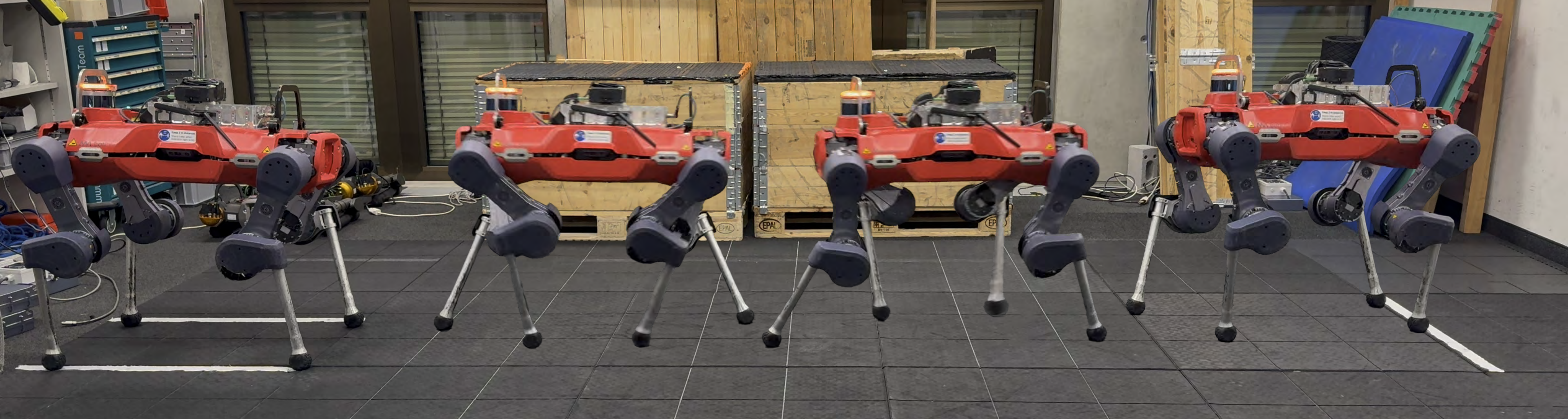}
            \caption{Real-world deployment with the proposed method.}
            \label{fig:results:sim2real:anymal:stills_pace}
        \end{subfigure}
    \end{minipage}%
    \hfill
    \begin{minipage}{0.49\textwidth}
        \begin{subfigure}{\linewidth}
            \centering
            \includegraphics[width=\linewidth]{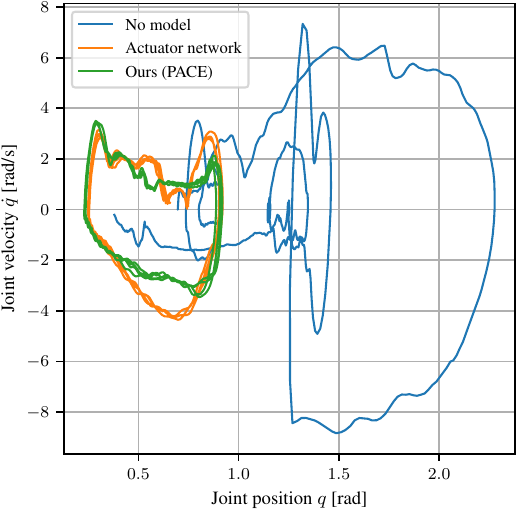}
            \caption{Phase portrait of the LF HFE joint over multiple gait cycles.}
            \label{fig:results:sim2real:anymal:phase_portraits}
        \end{subfigure}
    \end{minipage}

    \begin{subfigure}{\linewidth}
        \centering
        \includegraphics[width=\linewidth]{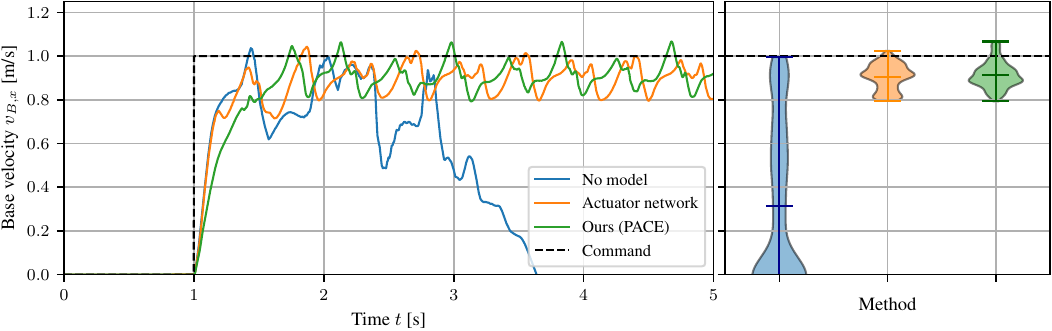}
        \caption{Forward velocity tracking. Left: commanded and measured velocities over time. Right: distribution of steady-state velocity errors across methods.}
        \label{fig:results:sim2real:anymal:forward_velocity_plot}
    \end{subfigure}\hfill

    \caption{Sim-to-real evaluation on \robot{ANYmal}. Panels (a–c) illustrate real-world deployments. Panel (d) shows the corresponding phase portraits. Panel (e) presents commanded and measured forward velocities (left) and steady-state velocity error distributions across methods (right).}
    \label{fig:results:sim2real:anymal}
\end{figure*}

\subsubsection{Full robot}\label{sec:experiments:full_robot}

At the system level we identify all parameters, including joint position biases and a global delay, with the base stationary and legs moving in air. We collect identification and validation sets on \robot{Tytan}, \robot{ANYmal}, and \robot{Minimal}. 
For \robot{Tytan}, identification uses sinusoidal joint targets at $P_\tau\!=\!\qty{60}{\newton\metre\per\radian}$, $D_\tau\!=\!\qty{2}{\newton\metre\second\per\radian}$; validation repeats with $P_\tau\!=\!\qty{145}{\newton\metre\per\radian}$, $D_\tau\!=\!\qty{5}{\newton\metre\second\per\radian}$. 
For \robot{Minimal}, we additionally validate on random joint steps (\qty{0.5}{\second} dwell). 
On \robot{ANYmal}, low-level gains are fixed; we therefore use the vendor defaults for both \ac{pace} and the actuator-network baseline. \rev{The actuator-network baseline is the ANYbotics LSTM actuator model deployed on our platform, rather than a model retrained on the suspended, contact-free chirp data used for \ac{pace} identification. Its proprietary training set is not available to us; however, the original actuator-network formulation was trained on contact-rich, high-frequency foot-trajectory data with measured torques~\cite{hwangbo2019learning}. We therefore interpret this comparison as an evaluation against a strong practical actuator-network baseline, not as a controlled equal-data comparison between model classes.}

\subsection{On-ground locomotion analysis}\label{sec:experiments:sim2real}
Having validated \ac{pace} under controlled conditions, we now assess its performance in locomotion scenarios under ground contact. 
These experiments evaluate full-robot deployment on hardware, testing whether the identified models enable robust tracking. All on-ground locomotion experiments use the actuator and delay parameters identified from in-air data (Section~\ref{sec:experiments:full_robot}) without further adaptation or re-identification. Ground-contact rollouts are not used for parameter estimation and serve only to evaluate sim-to-real transfer performance under realistic locomotion conditions.

We further analyze the two platforms \robot{Tytan} and \robot{ANYmal} on long-duration trials. We quantify locomotion efficiency and decompose the total energy consumption into contributions from locomotion, inverter switching losses, compute and sensors overhead.

Using the identified parameters, we train blind locomotion policies as in Section~\ref{sec:method:control} without dynamics randomization. Hyperparameters and robot specific parameters are listed in Table~\ref{tab:appendix:ppo_hyperparameters} and Table~\ref{tab:experiments:reward_scales} respectively. The same reward structure and scales are reused across \robot{ANYmal}, \robot{Tytan}, and \robot{Minimal}. 
One coefficient that differs is the energy penalty for Minimal, which is increased due to its lightweight, 3D-printed morphology. No task- or robot-specific retuning of tracking, collision or foot-touchdown terms was required.

\subsubsection{Tytan}\label{sec:experiments:sim2real:tytan}
We train a single policy and compare real vs.\ simulation by replaying the real commanded base-velocity trajectory in open loop. The robot is manually driven within a \qty{3}{\metre}\,$\times$\,\qty{3}{\metre} area; total duration is $\sim\!\qty{90}{\second}$ (SignalLogger buffer limit).

\subsubsection{ANYmal}\label{sec:experiments:sim2real:anymal}
We train three policies—(i) no model (URDF-only), (ii) actuator network~\cite{hwangbo2019learning}, and (iii) \ac{pace}—with identical rewards, no dynamics randomization, and the same environment. We compare step responses at \qty{1}{\metre\per\second} (forward/sideways over \qty{4}{\metre}) and yaw steps at \qty{2}{\radian\per\second}.

\subsubsection{Energetic running-track evaluations}\label{sec:experiments:running_track}

\begin{figure*}
    \centering

    \begin{subfigure}{\linewidth}
        \centering
        \includegraphics[width=\linewidth]{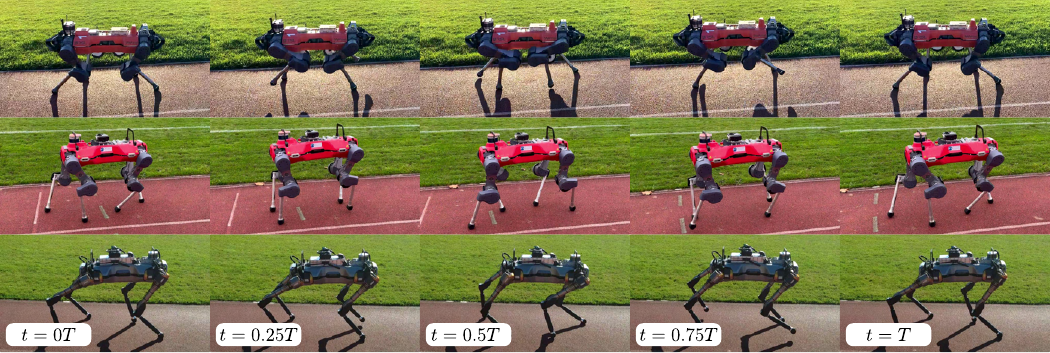}
        \caption{Representative stills from one step cycle during running-track evaluation.}
        \label{fig:results:running_track:stills}
    \end{subfigure}

        \begin{subfigure}{0.49\linewidth}
            \centering
            \includegraphics[width=\linewidth]{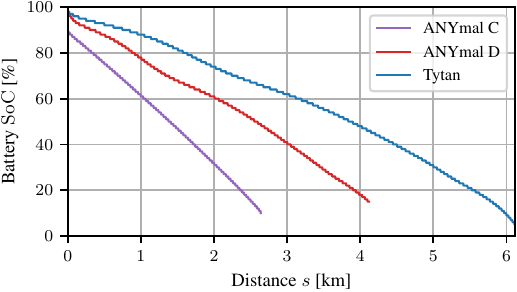}
            \caption{Battery state of charge (SoC) as a function of distance traveled.}
            \label{fig:results:running_track:soc_distance}
        \end{subfigure}\hfill
        \begin{subfigure}{0.49\linewidth}
            \centering
            \includegraphics[width=\linewidth]{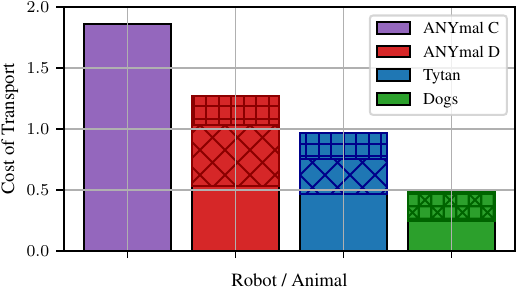}
            \caption{Decomposition of the Cost of Transport (CoT). }
            \label{fig:results:running_track:limit_cycles}
        \end{subfigure}
        
        \begin{subfigure}{0.49\linewidth}
            \centering
            \includegraphics[width=\linewidth]{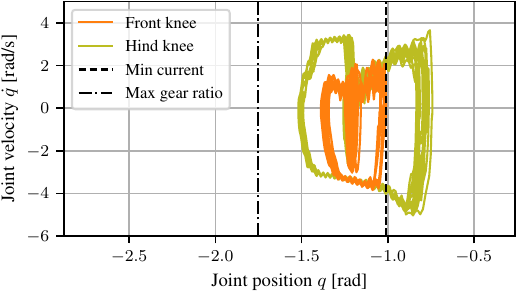}
            \caption{Limit cycle of the \robot{Tytan} knee joint over a \SI{10}{\second} window.}
            \label{fig:results:running_track:cot}
        \end{subfigure}\hfill
        \begin{subfigure}{0.49\linewidth}
            \centering
            \includegraphics[width=\linewidth]{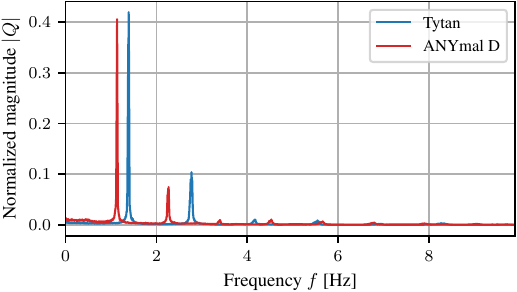}
            \caption{Frequency spectrum of the LF HFE joint during steady locomotion.}
            \label{fig:results:running_track:frequency_spec}
        \end{subfigure}

    \caption{Running-track evaluations across robotic platforms. 
    (a) Representative stills from one step cycle. 
    (b) Battery consumption over distance (SoC).
    (c) Cost of transport (CoT) with contributions from electronics (CoE, diagonal hatching) and drives' inverter switching (CoD, horizontal hatching). 
    (d) Limit cycle of the \robot{Tytan} knee joint over a \SI{10}{\second} window. 
    (e) Frequency spectrum of the LF HFE joint during steady locomotion.}
    \label{fig:results:running_track}
\end{figure*}

\begin{table*}
\small\sf\centering
\caption{Running-track performance (per full battery charge). \robot{ANYmal}~C and \robot{AoPS} are from~\cite{bjelonic2023learning}.}
\label{tab:energy_efficiency}
\begin{tabular}{c|c|c||c|c}
\toprule
& \robot{ANYmal}~C & \robot{AoPS} & \robot{ANYmal}~D & \robot{Tytan} \\
\hline\hline
Rounds [--] & \num{6.6} & \num{7.5} & \num{10.3} & \num{15.25} \\
Distance $s$ [\si{\meter}] & \num{2640} & \num{3000} & \num{4120} & \num{6100} \\
Initial SoC [\si{\percent}] & \num{89} & \num{92} & \num{95} & \num{98} \\
Final SoC [\si{\percent}] & \num{10} & \num{11} & \num{15} & \num{5} \\
Time $t$ [\si{\minute}] & \num{59} & \num{68} & \num{82} & \num{104} \\
Commanded $\hat{v}_{x,B}$ [\si{\meter \per \second}] & \num{1.0} & \num{1.0} & \num{1.0} & \num{1.0} \\
Average $\bar{v}_{x,B}$ [\si{\meter \per \second}] & \num{0.740} & \num{0.735} & \num{0.850} & \num{0.978} \\
Efficiency $\eta$ [\si{\percent}] & \num{100} & \num{111} & \num{132} & \num{196} \\
Ambient Temp. $\Theta$ [\si{\celsius}] & \num{26} & \num{31} & \num{10} & \num{18} \\
Avg.\ power $P_\mathrm{full}$ [\si{\watt}] & \num{723} & \num{646} & \num{556} & \num{504} \\
\bottomrule
\end{tabular}
\end{table*}

\begin{table}
\small\sf\centering
\caption{CoT decomposition on the track.}
\label{tab:results:running_track:cot}
\begin{tabular}{c|c|c|c||c}
\toprule
& CoE & CoD & CoL & CoT \\
\hline\hline
\robot{ANYmal} C & (\num{0.50}) & (\num{0.24}) & \num{1.12} & \num{1.86} \\
\robot{ANYmal} D & \num{0.50} & \num{0.24} & \num{0.53} & \num{1.27} \\
\robot{Tytan}    & \num{0.29} & \num{0.21} & \num{0.47} & \num{0.97} \\
Dogs & \multicolumn{2}{c|}{\num{0.25}} & \num{0.23} & \num{0.48} \\
\bottomrule
\end{tabular}
\end{table}

\rev{With dynamics fitted by \ac{pace}, the policies evaluated here do not require dynamics randomization to transfer to hardware, allowing the controller to specialize to the identified nominal machine dynamics. We do not claim that this eliminates the value of domain randomization or online adaptation; rather, these methods are complementary and may further improve robustness under unmodeled environmental or hardware variations.} We emphasize that this outcome is not unique to PACE (e.g.,~\cite{miller2025high}), but arises here from explicit actuator-level identification using a compact, physically interpretable parameterization. We therefore perform long-duration endurance tests on a standard \qty{400}{\metre} track, comparing to our prior actuator-network results~\cite{bjelonic2023learning}. The same \robot{Tytan} and \robot{ANYmal} policies from Secs.~\ref{sec:experiments:sim2real:tytan} and \ref{sec:experiments:sim2real:anymal} are used. 
All policies are also capable of traversing standard stairs (cf.\ Figure~\ref{fig:approach_overview}). 

Both platforms use the same battery type as in~\cite{bjelonic2023learning}, with capacity $E_B=\qty{907.2}{\watt\hour}$. 
We decompose the total cost of transport (CoT) additively as
\begin{equation}
\mathrm{CoT} = \mathrm{CoE} + \mathrm{CoD} + \mathrm{CoL},
\end{equation}
where each term captures a disjoint contribution. 
\textbf{CoE} accounts for controller-independent electronics power (compute and sensing), measured with the robot powered but drives disabled.
\textbf{CoD} captures inverter and drive overheads, measured with drives enabled but zero torque.
\textbf{CoL} contains only motion-dependent losses during locomotion, including mechanical dissipation and electrical losses such as Joule heating.
By construction, CoE and CoD are largely independent of gait mechanics, while CoL reflects the portion directly influenced by modeling accuracy and control.

To estimate resting powers, we repeat full-charge discharge tests with (i) robot on a crane, drives enabled but zero current target (\emph{rest}), and (ii) robot on ground, drives off (\emph{off}). With measured durations $t_i$ and \ac{soc} windows $(\mathrm{SoC}_i^\text{max}-\mathrm{SoC}_i^\text{min})$, the average powers are
\begin{align}
    P_\text{rest} &= \frac{E_B\big(\mathrm{SoC}_\text{rest}^\text{max}-\mathrm{SoC}_\text{rest}^\text{min}\big)}{t_\text{rest}},\\
    P_\text{off}  &= \frac{E_B\big(\mathrm{SoC}_\text{off}^\text{max}-\mathrm{SoC}_\text{off}^\text{min}\big)}{t_\text{off}}.
\end{align}
For a locomotion experiment of duration $t_\text{track}$ and distance $\Delta s$, the total CoT is
\begin{align}
    \mathrm{CoT} \;=\; \frac{E_B\big(\mathrm{SoC}_\text{track}^\text{max}-\mathrm{SoC}_\text{track}^\text{min}\big)}{m g\,\Delta s}.
\end{align}
Electronics and drives contributions are
\begin{align}
    \mathrm{CoE} &= \frac{P_\text{off}\,t_\text{track}}{m g\,\Delta s}, \qquad
    \mathrm{CoD} \;=\; \frac{\big(P_\text{rest}-P_\text{off}\big)\,t_\text{track}}{m g\,\Delta s},
\end{align}
and the locomotion term follows as
\begin{align}
    \mathrm{CoL} \;=\; \mathrm{CoT} - \frac{P_\text{rest}\,t_\text{track}}{m g\,\Delta s}.
\end{align}

\section{Results}\label{sec:results}

We report identification progress, fitted parameters, and sim2real performance across \robot{Tytan}, \robot{ANYmal}, and \robot{Minimal}. Optimization traces (\ac{cmaes} scores per iteration) are shown in Figure~\ref{fig:results:fitting_scores}; final parameter sets appear in Table~\ref{tab:appendix:pace_params}. For \robot{ANYmal}, the same fitted set can be used across units, though hardware wear, varying firmware and changing simulator may introduce small residual gaps.

Across robots, same-type joints fit to similar armature $I_a$ and damping $d$. Both \robot{ANYmal} and \robot{Tytan} identify a global delay of \qty{7.5}{\milli\second}. \robot{ANYmal} shows the largest fitted damping (order \qty{5}{\newton\meter\second\per\radian}); pseudo-direct variants (\robot{Tytan}, \robot{Minimal}) fit substantially lower $d$. Despite higher damping on \robot{ANYmal}, fitted Coulomb-like friction is of similar order to \robot{Tytan}. With $N\!=\!4096$ environments, end-to-end optimization for each platform converges within \qtyrange{10}{24}{\hour}, depending on trajectory length, GPU throughput, and parameter bounds.

\subsection{In-air evaluation and validation}\label{sec:results:model_analysis}

\subsubsection{Single drive}\label{sec:results:single_drive}

\paragraph{Current loop validation.}

From current-chirp data (\qtyrange{1}{1250}{\hertz}, \qty{2}{\ampere}, \qty{25}{\second}) we estimate the motor inner-loop transfer $H_i(s)=\tau_m/\hat{\tau}_m$. Measured Bode plots (Figure~\ref{fig:appendix:current_tracking}) match the analytical model up to $>\!\qty{100}{\hertz}$. Magnitude stays near \qty{0}{\decibel} (unity gain) with $\le\!\qty{5}{\degree}$ phase lag up to \qty{25}{\hertz}; the $-1\,\si{\decibel}$ point occurs near \qty{50}{\hertz}, and we observe a control bandwidth of $\approx\!\qty{346}{\hertz}$ or \qty{0.58}{\percent} of the pulse-width modulation frequency. A dead time of $\approx\!\qty{400}{\micro\second}$ is inferred from the phase slope. Above $\sim\!\qty{350}{\hertz}$, signal-to-noise ratio limits the estimate. 

\begin{figure*}
    \centering
    \includegraphics[width=\linewidth]{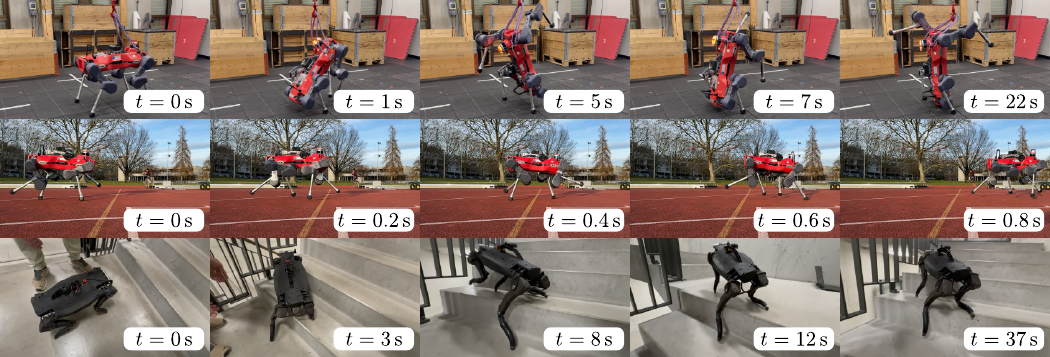}
    \caption{Dynamic-limit demonstrations. Top: two-legged balance with \robot{ANYmal}. Center: running with \robot{ANYmal}. Bottom: stair climbing with \robot{Minimal}.}
    \label{fig:results:dynamic_limits}
\end{figure*}

\begin{figure}
    \centering
    \includegraphics[width=\linewidth]{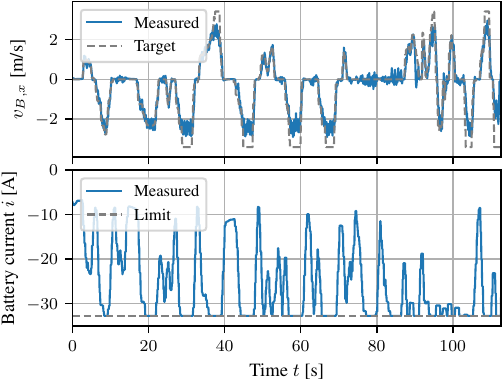}
    \caption{Running experiments with \robot{ANYmal}. Top: commanded forward velocity versus measured base velocity (state estimation). Bottom: battery current profile during the same run, including measured current and the saturated peak limit of \qty{32}{\ampere} imposed by the battery.}
    \label{fig:results:anymal_running}
\end{figure}

\paragraph{Mechanical loop identification.}
We fit $\{I_a,d,\tau_f\}$ from position-chirp data (\qtyrange{0.1}{10}{\hertz}, \qty{2.5}{\kilo\hertz} logging) over five load cases and three PD configurations (Section~\ref{sec:method:single_drive}). In the no-compensation firmware mode, fitted $I_a$ agrees with analytic targets within \qty{2}{\percent} (rotor), \qty{6}{\percent} (interface), and \qty{15}{\percent} (added mass at radius); across PD settings, the standard deviation is \qty{0.67}{\percent}, \qty{3.7}{\percent}, and \qty{14}{\percent}, respectively (Figure~\ref{fig:results:single_drive}). One high-gain/high-mass trial failed mechanically (the lever broke) and is excluded. Parameter variations remain small across the locomotion-relevant operating range and load-cases, with noticeable deviations only near extreme conditions close to material limits, which are outside the operating regime considered in this work.

With firmware compensations enabled, all fits exhibit an approximately constant offset $I_\mathrm{comp}\!=\!\qty{8.1e-3}{\kilogram\metre\squared}$ (virtual inertia, cf. Figure~\ref{fig:appendix:compansations_on}). 

Assuming unit inner-loop gain, the outer-loop single-joint transfer from $\hat{q}$ to $q$ matches the second-order model in Equation~\ref{eq:method:transfer_function}. For the high-load, locomotion-gain case ($P_\tau\!=\!\qty{60}{\newton\meter\per\radian}$, $D_\tau\!=\!\qty{2}{\newton\meter\second\per\radian}$), the dominant complex conjugate yields a gain drop near \qty{4.35}{\hertz} and $\approx\!-14\,\si{dB}$ at \qty{10}{\hertz} (cf. Figure~\ref{fig:appendix:bode_plot}). The dominant pole pair lies within the frequency range spanned by the excitation trajectories.

\subsubsection{Full robot} \label{sec:results:full_robot}

Representative in-air trajectory overlays for \robot{Tytan} show near-overlap between real measurements and simulated replay using fitted parameters (Figure~\ref{fig:results:tytan_hfe_trajectory_evaluation}). Validation at different gains ($P_\tau\!=\!\qty{145}{\newton\meter\per\radian}, D_\tau\!=\!\qty{5}{\newton\meter\second\per\radian}$) preserves the match (Figure~\ref{fig:results:tytan_hfe_trajectory_validation}).

A comparison against the single-drive baseline highlights a clear discrepancy. 
For the LF–HFE joint of \robot{Tytan}, the fitted inertia is $I_a=\qty{10.6e-2}{\kilogram\metre\squared}$, 
approximately four times higher than the expected $I_a\approx\qty{2.5e-2}{\kilogram\metre\squared}$ from rotor and compensation alone, while the fitted damping $d=\qty{0.17}{\newton\meter\second\per\radian}$ remains consistent with expectation (cf.\ Table~\ref{tab:appendix:pace_params}).

On \robot{Minimal}, both chirps and random joint steps are reproduced closely; the URDF-only model deviates substantially (Appendix: Figure~\ref{fig:appendix:minimal_chirp}, Figure~\ref{fig:appendix:minimal_random}).

\paragraph{Comparison with the actuator network on ANYmal} \label{sec:results:actuator_network}

Figure~\ref{fig:main_figure} compares in-air behavior for \robot{ANYmal} using: (i) no model (URDF-only), (ii) the actuator network~\cite{hwangbo2019learning}, and (iii) \ac{pace}. The trace overlays (stills) qualitatively match only for the latter two; the no-model case diverges. The delta phase portraits of LF–HAA show smallest $(\Delta q,\Delta\dot{q})$ with \ac{pace}, while the actuator network exhibits more variance and a joint position bias (\(\sim\!\qty{5}{\degree}\)). The deltas are calculated as $\Delta q_i = q^\mathrm{sim}_i - q^\mathrm{real}_i$ and $\Delta \dot{q}_i = \dot{q}^\mathrm{sim}_i - \dot{q}^\mathrm{real}_i$. The closer a trajectory to (0,0), the smaller the reality gap.

A zoomed HFE time window appears in Figure~\ref{fig:results:anymal_air_hfe_trajectory}. Data usage differs: our approach relied on \(\sim\!\qty{20}{\second}\) of in-air data; the 2019 actuator network used \(\sim\!\qty{4}{\minute}\) of torqued data, using more complex data acquisition techniques, and the deployed vendor LSTM was likely trained on a larger dataset.

\subsection{On-ground locomotion analysis}\label{sec:results:sim2real}

\subsubsection{Tytan}

Figure~\ref{fig:results:sim2real:tytan} juxtaposes real vs.\ simulated sequences for an open-loop replay of commanded base velocities. Both absolute and relative motion match closely (cf. supplementary video). Linear and yaw-rate tracking align between real (blue) and fitted simulation (orange), with targets in dashed black. Measured vs.\ commanded motor torques on the LF leg also match well; small deviations appear at KFE during liftoff.

\subsubsection{ANYmal}

Forward walking with URDF-only fails (caught by crane), whereas both actuator network and \ac{pace} succeed (Figure~\ref{fig:results:sim2real:anymal}). During steady state, both achieve a mean \(\approx\!\qty{0.85}{\meter\per\second}\) (violin plot), with similar limit cycles; the no-model case diverges by \(\sim\!\qty{3}{\second}\). Sideways and yaw tests are provided in the Appendix (Figure~\ref{fig:results:anymal:appendix}); the URDF-only policy can execute yaw steps at \qty{2}{\radian\per\second} but remains unreliable for \qty{1}{\meter\per\second} forward and sideways walking.

\subsubsection{Energetic running-track evaluations}\label{sec:results:running_track}

We replicate the \qty{400}{\meter} track protocol of~\cite{bjelonic2023learning} and extend it with \robot{ANYmal}~D and \robot{Tytan}. Summary statistics are listed in Table~\ref{tab:energy_efficiency}. State-of-charge vs.\ distance, CoT decomposition, frequency spectra, and a knee limit-cycle analysis are in Figure~\ref{fig:results:running_track}.

We decompose the cost of transport (CoT) into electronics (CoE), drives (CoD), and locomotion (CoL) using the procedures in Section~\ref{sec:experiments:running_track}. Table~\ref{tab:results:running_track:cot} summarizes the contributions. For \robot{ANYmal}~D, \ac{cot} reduces to \num{1.27} (vs.\ \num{1.86} for \robot{ANYmal}~C); \robot{Tytan} reaches \num{0.97}. In both platforms, less than half of the energy is attributed to CoL. Biological dogs reach similar distribution at half the \ac{cot}~\cite{bryce2017comparative} at \qty{30}{\kilogram} weight compared to \robot{Tytan}.

\subsubsection{Dynamic-limit demonstrations} \label{sec:results:dynamic_limit}

Figure~\ref{fig:results:dynamic_limits} shows three policies trained in fitted simulation and deployed zero-shot. On \robot{ANYmal}, we demonstrate two-legged balancing (orientation tracking) and running. In these running trials, simulation attained \(\approx\!\qty{4}{\meter\per\second}\); hardware peaked near \qty{3}{\meter\per\second}, limited by battery current (\qty{32}{\ampere} threshold; Figure~\ref{fig:results:anymal_running}). The running speed is presented as contextual comparison and not claimed as a new overall record. On \robot{Minimal} (\qty{4}{\kilogram}), the policy climbs standard \qty{18}{\centi\meter} stairs continuously (20 steps in \qty{46}{\second}).

\section{Discussion} \label{sec:discussion}

\textbf{Inner loop as near–unit torque source.}
The electrical inner loop tracks torque at high bandwidth (measured \( \approx \) \qty{346}{\hertz}), so within the policy’s frequency range it can be treated as a near–unit-gain source (Section~\ref{sec:results:single_drive}). With more advanced control, bandwidths up to \( \sim \)\qty{3.5}{\kilo\hertz} are feasible~\cite{springob2002high}.

\textbf{Physicality of fitted parameters and \emph{virtual} inertia.}
On a single drive, \ac{pace} recovers output inertias that match analytic expectations, and the drive behaves closely like a second–order LTI system. Because the mechanical bandwidth falls in the excitation range, $I_a$ and $d$ can be reliably identified and directly relate to the dynamics shaped by the PD gains (cf.\ Section~\ref{sec:method:data_collection:pd_gains}). When firmware compensations are enabled, we observe an additive, load–independent \emph{virtual} inertia that shifts the effective dynamics (Section~\ref{sec:results:single_drive}); identification and deployment should therefore use identical firmware modes.

\textbf{Interpreting \(I_a\) and \(d\).}
Across abstraction levels, the fitted armature term \(I_a\) aggregates rotor inertia, compensation effects, and inaccuracies in link and load modeling, while the fitted damping term \(d\) captures contributions from the motor, gearbox, and compensations. Both scale with the ratio \(k_i/k_{i,\mathrm{real}}\), and are handled end–to–end by \ac{pace}. 
The unexpectedly larger full-robot \(I_a\) (about fourfold relative to the single-drive baseline) is consistent with three effects: (i) underestimation of link inertia in CAD, particularly in the thigh, (ii) additional apparent inertia introduced by compensation, and (iii) rotor inertia. 
A diagnostic single-leg-segment analysis confirms this explanation (Appendix~\ref{sec:appendix:link_inertia_estimation}), but it is not central to our contribution.

\textbf{Generalization across gains and trajectories.}
Fitted simulators reproduce in–air trajectories at unseen PD gains and for unseen trajectories with higher fidelity than the actuator–network baseline, indicating a consistent physical model rather than an overfit function approximation (cf.\ Figure~\ref{fig:results:tytan_hfe_trajectory_evaluation}, Figure~\ref{fig:results:anymal_air_hfe_trajectory} and Figure~\ref{fig:appendix:minimal_sim2real}).

\textbf{In–air data suffices for contact tasks.}
Identifying joint–space dynamics (Section~\ref{sec:method:pace}) is sufficient for zero–shot locomotion. Our results do not contradict prior actuator networks~\cite{hwangbo2019learning}, but reflect a different operating regime. We deliberately identify dynamics from contact-free, in-air excitation. In this regime, the mechanical loop dominates and the actuator is well approximated by a low-order LTI model; remaining nonlinearities are either outside the excitation range or absorbed into the fitted parameters.

\textbf{Zero–model failure and the role of speed.}
Policies trained without actuator models can fail even at low speeds (Section~\ref{sec:results:sim2real}); covering the resulting gap would require heavy dynamics randomization, whose scope grows rapidly with step frequency (Figure~\ref{fig:appendix:domain_rand}). Thus, achievable speed (step frequency) is a sensitive proxy for the reality gap.

\textbf{Sim2real tracking and dynamic behaviors.}
Open–loop base–velocity replays match between real and simulation (Figure~\ref{fig:results:sim2real:tytan}), and \ac{pace} is competitive with the actuator network on \robot{ANYmal} (Figure~\ref{fig:results:sim2real:anymal}). The same modeling enabled two dynamic behaviors on \robot{ANYmal}: two-legged balancing and high-speed running up to the platform's electrical limit. It also enabled continuous stair-climbing on \robot{Minimal}. The \qty{3}{\meter\per\second} ceiling on \robot{ANYmal} is set by battery current limits (\qty{32}{A}, Figure~\ref{fig:results:anymal_running}). Notably, \ac{pace} uses only \(\sim\)\qty{20}{s} of encoder–only in–air data per robot, versus minutes of torque–instrumented data for actuator networks—broadening applicability to systems without torque sensors.

\textbf{Energy shaping and straight–leg gaits.}
With compensation and \(k_i\) drift folded into \(d\), the energy reward (Section~\ref{sec:method:rewards}) penalizes dissipations that correlate with track–side CoT reductions (Table~\ref{tab:results:running_track:cot}). \rev{In the evaluated setting, precise simulator fitting removes the need for dynamics randomization during policy training, enabling \robot{ANYmal} to adopt straighter-knee gaits in the identified nominal dynamics.} This reduces the total CoT by \(\sim\)\qty{32}{\percent} and halves locomotion losses. Despite \robot{Tytan}'s efficient drives and low leg inertia, its inability to fully straighten the knee keeps its locomotion cost only \(\sim\)\qty{11}{\percent} lower than \robot{ANYmal}'s in this regime, underscoring the energetic importance of full knee extension on flat terrain (Table~\ref{tab:results:running_track:cot}).

\textbf{Where the watts go.}
For both \robot{ANYmal} and \robot{Tytan}, less than half of the energy budget is spent on locomotion itself; electronics and inverter switching account for the remainder. Algorithmic efficiency alone cannot close the full \ac{cot} gap to biology. We need to build robots based on more efficient sensors, computers, and power electronics.

\textbf{Generalization across robots.}
The same parameterization
\(\{\mathbf{I}_a,\mathbf{d},\boldsymbol{\tau}_f,\tilde{\mathbf{q}}_b,T_d\}\)
transfers from open (\robot{Tytan}, \robot{Minimal}) to closed (\robot{ANYmal}) platforms—and beyond—suggesting a broadly useful basis.

\textbf{Future extensions.}
\begin{itemize}
    \item Hybrid CMA–ES with local gradient refinements in differentiable physics.
    \item Contact–parameter refinement when foot–force sensing or plates are available.
    \item Lightweight online adaptation (global scale on \(\{I_a,d\}\)) to track wear/temperature over time~\cite{lee2020learning}.
\end{itemize}

\section{Conclusion}
We presented \emph{\acf{pace}}, an alignment of joint-space dynamics that enables zero-shot sim2real locomotion on the evaluated platforms while eliminating broad actuator-dynamics randomization during policy training. The key idea is to fit a parameterization—per-joint armature, viscous damping, Coulomb friction, joint bias, and a global delay—directly using \(\sim\)\qty{20}{s} of in-air, encoder-only trajectories and massively parallel evolutionary search.

Across three platforms (\robot{Tytan}, \robot{ANYmal}, \robot{Minimal}), the fitted simulators reproduce in-air joint trajectories with near overlap and generalize across PD gains and trajectories; on \robot{ANYmal} they close the gap where URDF-only models fail and match actuator-network fidelity while requiring less data and no torque sensors. The fitted dynamics translates to the ground: blind locomotion policies train in fitted simulation and deploy zero-shot, yielding accurate base-velocity tracking and endurance improvements on a \qty{400}{m} track (e.g., \robot{ANYmal}~D: CoT \num{1.27} and \qty{4.12}{\kilo \meter}; \robot{Tytan}: CoT \num{0.97} and \qty{6.10}{\kilo \meter}). Dynamic-limit demonstrations (two-legged balance, \(\sim\)\qty{3}{m/s} running bounded by a \qty{32}{A} battery limit, and stair climbing on a \qty{4}{\kilogram} robot) further indicate that residual constraints are hardware–power and sensing limited rather than model limited.

The approach is data- and compute-efficient (\qty{20}{\second} trajectories, single-GPU, \qtyrange{10}{24}{h} per robot with \(N\!=\!4096\) environments) and applies to both open and closed platforms, provided firmware compensation modes and filters are consistent between identification and deployment. Current limitations include finite excitation bandwidth on suspended setups and temperature/aging dependencies that shift effective parameters. Our approach fails if any of the assumptions from Section~\ref{sec:method:remarks} are violated, which fully online and model-free approaches might be able to catch.

Future work will focus on (i) accurately identifying \emph{electrical} constraints—bus-voltage (Appendix~\ref{sec:appendix:pmsm_limit}) and current limits (Figure~\ref{fig:results:anymal_running}), and inverter switching behavior—and embedding them into the fit, and (ii) explicitly modeling \emph{compliance} (joint/foot stiffness, link flexibilities, series elasticity) within the optimization. Together, these additions should enable higher step-frequency gaits that deliberately exploit hardware dynamics rather than fight them. As simulators expose richer effects, we will also extend the parameterization beyond acceleration to include higher-order motion terms—jerk \(\dddot{q}\), snap \(q^{(4)}\), crackle \(q^{(5)}\), and pop \(q^{(6)}\)—to better capture bandwidth limits, electro-magnetic effects, mitigate wear, and shape smooth, high-frequency locomotion.

\noindent\textbf{Takeaway.} A small, physically meaningful parameter set ($4n{+}1$), identified from in-air experiments using joint encoders, \rev{was sufficient in our experiments to eliminate broad dynamics randomization} for blind quadrupedal locomotion and to translate simulation capabilities to hardware in a single shot.


\begin{an}
ORCIDs \\
\noindent Filip Bjelonic \href{https://orcid.org/0000-0002-4890-3132}{0000-0002-4890-3132} \\
\noindent Fabian Tischhauser \href{https://orcid.org/0009-0009-8821-3994}{0009-0009-8821-3994} \\
\noindent Marco Hutter \href{https://orcid.org/0000-0002-4285-4990}{0000-0002-4285-4990}
\end{an}

\begin{acks}
The authors would like to thank the RSL Learning Group for many insightful discussions throughout this work. We are grateful to Konrad and Matthias for valuable discussions on electronics. We also thank Zichong, Stephan, Efe, Yuntao, René, Clemens, Ryo, Alexander, Fabio for employing and extending our approach in their own research. Finally, we acknowledge the use of OpenAI’s GPT-5, which assisted in refining the manuscript language. All technical content, analyses, and citations were generated and verified by the authors. We note that large language models may exhibit biases, errors, or omissions, and we take full responsibility for the accuracy and appropriateness of the manuscript.
\end{acks}

\begin{AuthorContribution}
The authors confirm contribution to the paper as follows: 
study conception and design: F. Bjelonic, F. Tischhauser, M. Hutter; 
data collection: F. Bjelonic; 
analysis and interpretation of results: F. Bjelonic; 
mechanical experimental setups and electronics support: F. Tischhauser; 
manuscript preparation: F. Bjelonic, M. Hutter. 
M. Hutter provided funding, supervision and critical feedback. 
All authors reviewed the results and approved the final version of the manuscript.
\end{AuthorContribution}

\begin{dci}
    The authors declare that there are no potential conflicts of interest with respect to the research, authorship, or publication of this article.
\end{dci}

\begin{funding}
    This work was supported by the European Union’s Horizon Europe Framework Programme (Grant Agreement No. 101070596).
\end{funding}

\begin{das}
    The datasets supporting the findings of this study are available at \url{https://doi.org/10.3929/ethz-c-000783505}, and the source code is publicly available at \url{https://github.com/leggedrobotics/pace-sim2real}.
\end{das}


\bibliographystyle{agsm}
\bibliography{references}

\appendix

\begin{figure}
    \centering
    \includegraphics[width=\linewidth]{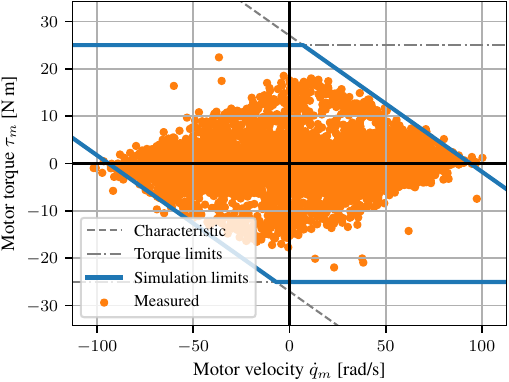}
    \caption{Idealized PMSM saturation envelope for the \robot{Tytan} hip motor, showing motor velocity $\omega$ (x-axis) versus torque $\tau$ (y-axis). The solid blue curve indicates the enforced torque limits in simulation: commanded torques inside the envelope are applied directly, while those outside are saturated to the boundary. This boundary is determined by two constraints: the maximum torque limit (dot-dashed gray), representing effects such as magnet demagnetization, and the back-EMF–limited torque (dashed gray). Orange dots mark experimental measurements from Section~\ref{sec:method:single_drive}.}
    \label{fig:torque_velocity_saturation_model}
\end{figure}

\begin{figure*}
    \centering
    \includegraphics[width=\linewidth]{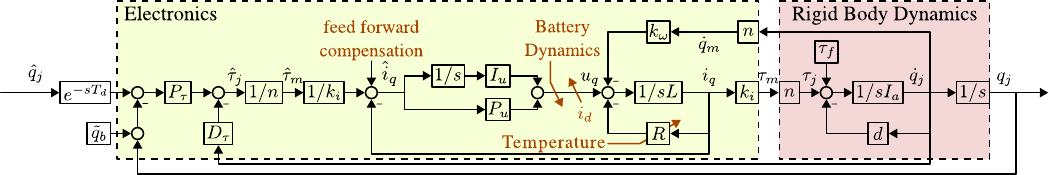}
    \caption{Simplified block diagram of the motor‐driven joint actuator, illustrating the electronics (left, yellow) and rigid‐body dynamics (right, red) subsystems. In the electronics path, a PD position controller with gains $P_\tau$ and $D_\tau$ drives a PI current loop—comprised of integrator $I_u$ and feedforward $P_u$—to regulate the torque producing quadrature current $i_q$. This path is subject to disturbances from the load current, nonlinear battery dynamics, temperature variations, and unknown feed-forward compensations. In the mechanical path, $i_q$ is converted to torque through the motor constant $k_i$ and gear ratio $n$, then transmitted to the inertia $I_a$, friction torque $\tau_f$, and viscous transmission damping $d$, yielding joint acceleration $\ddot q$ and position $q$. The two subsystems interact via the back‐EMF constant $k_\omega$.}
    \label{fig:background:blockdiagram_laplace}
\end{figure*}

\begin{sidewaystable}
\small\sf\centering
\caption{All PACE parameters found for \robot{Tytan}, \robot{ANYmal}, and \robot{Minimal}}
\label{tab:appendix:pace_params}
\begin{tabular}{c|c|c|c|c|c|c|c|c|c|c|c|c} 
\toprule
  Joint & LF HAA & LF HFE & LF KFE & RF HAA & RF HFE & RF KFE & LH HAA & LH HFE & LH KFE & RH HAA & RH HFE & RH KFE \\
 \hline
 \hline
 \multicolumn{13}{c}{Armature $I_a$ [\SI{e-3}{\kilogram \meter \squared}]}  \\
 \hline
 \robot{Tytan}   & \num{140} & \num{106} & \num{3.3} & \num{120} & \num{120} & \num{3.4} & \num{140} & \num{110} & \num{3.6} & \num{140} & \num{110} & \num{3.5} \\
 \robot{ANYmal}   & \num{76} & \num{76} & \num{67} & \num{74} & \num{77} & \num{67} & \num{89} & \num{51} & \num{64} & \num{79} & \num{39} & \num{51} \\
 \robot{Minimal} & \num{0.050} & \num{0.025} & \num{0.026} & \num{0.056} & \num{0.025} & \num{0.025} & \num{0.049} & \num{0.023} & \num{0.026} & \num{0.052} & \num{0.028} & \num{0.025} \\
 \hline
 \multicolumn{13}{c}{Damping $d$ [\SI{}{\newton \meter \second \per \radian}]}  \\
 \hline
 \robot{Tytan} & \num{1.7} & \num{0.17} & \num{2.1} & \num{1.7} & \num{0.22} & \num{2.3} & \num{0.50} & \num{0.63} & \num{3.9} & \num{0.70} & \num{0.83} & \num{2.3} \\
 \robot{ANYmal} & \num{4.9} & \num{4.4} & \num{5.2} & \num{4.7} & \num{4.3} & \num{5.3} & \num{4.9} & \num{4.9} & \num{5.4} & \num{5.1} & \num{5.1} & \num{5.5} \\
 \robot{Minimal} & \num{0.0} & \num{0.0} & \num{0.092} & \num{0.1911} & \num{0.031} & \num{0.030} & \num{0.0} & \num{0.0} & \num{0.066} & \num{0.0} & \num{0.0} & \num{0.030} \\
  \hline
 \multicolumn{13}{c}{Friction $\tau_f$ [--]}  \\
 \hline
 \robot{Tytan} & \num{0.0093} & \num{0.044} & \num{0.00025} & \num{0.0036} & \num{0.036} & \num{0.0015} & \num{0.0} & \num{0.031} & \num{0.0010} & \num{0.0} & \num{0.035} & \num{0.00050} \\
 \robot{ANYmal} & \num{0.0054} & \num{0.021} & \num{0.028} & \num{0.0035} & \num{0.027} & \num{0.036} & \num{0.0032} & \num{0.024} & \num{0.040} & \num{0.0029} & \num{0.013} & \num{0.045} \\
 \robot{Minimal} & \num{0.038} & \num{0.084} & \num{0.34} & \num{0.033} & \num{0.070} & \num{0.32} & \num{0.051} & \num{0.091} & \num{0.31} & \num{0.048} & \num{0.075} & \num{0.39} \\
 \hline
 \multicolumn{13}{c}{Joint Position Bias $\tilde{q}_b$ [\SI{}{\radian}]}  \\
 \hline
 \robot{Tytan} & \num{0.0017} & \num{-0.011} & \num{-0.028} & \num{-0.0029} & \num{-0.012} & \num{-0.026} & \num{-0.0011} & \num{-0.017} & \num{-0.026} & \num{-0.00070} & \num{-0.0148} & \num{-0.0275} \\
 \robot{ANYmal} & \num{0.022} & \num{0.0057} & \num{-0.003} & \num{0.011} & \num{-0.0072} & \num{0.0094} & \num{-0.012} & \num{-0.0013} & \num{-0.0095} & \num{-0.016} & \num{0.0043} & \num{0.0045} \\
 \robot{Minimal} & \num{-0.018} & \num{-0.0030} & \num{-0.0017} & \num{-0.020} & \num{0.010} & \num{-0.018} & \num{-0.00040} & \num{0.0012} & \num{0.0027} & \num{0.010} & \num{-0.0065} & \num{0.0188} \\
 \hline
 \multicolumn{13}{c}{Delay $T_d$ [\SI{}{\milli \second}]}  \\
 \hline
 \robot{Tytan}   &  \multicolumn{12}{l}{\num{7.5}} \\
 \robot{ANYmal}   & \multicolumn{12}{l}{\num{7.5}} \\
 \robot{Minimal} &  \multicolumn{12}{l}{\num{0.0}} \\
 \bottomrule
\end{tabular}
\end{sidewaystable}

\begin{figure}
    \centering
    \includegraphics[width=\linewidth]{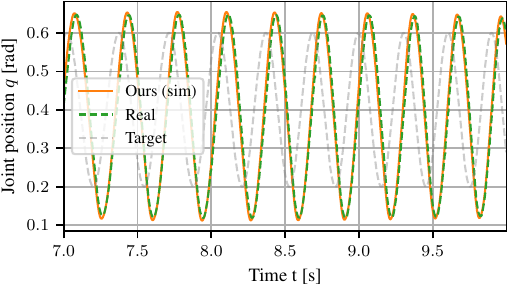}
    \caption{Validation of \robot{Tytan} on the HFE joint using proportional–derivative (PD) gains of 145/5. Shown are commanded trajectories, real measurements and the response under the near-optimal identified model.}
    \label{fig:results:tytan_hfe_trajectory_validation}
\end{figure}

\begin{figure}
    \centering
    \includegraphics[width=\linewidth]{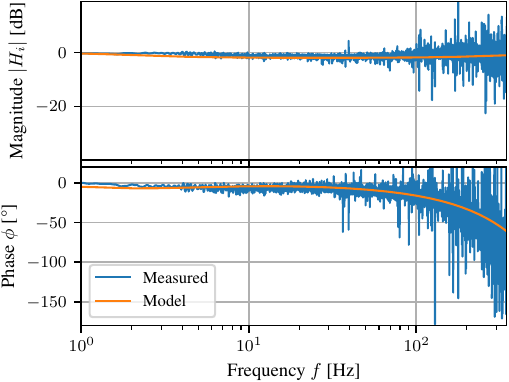}
    \caption{Current tracking performance for the inner-loop transfer function $H_i : \hat{\tau}_m \rightarrow \tau_m$, mapping commanded motor torque $\hat{\tau}_m$ to measured torque $\tau_m$.}
    \label{fig:appendix:current_tracking}
\end{figure}

\begin{figure}
    \centering
    \includegraphics[width=\linewidth]{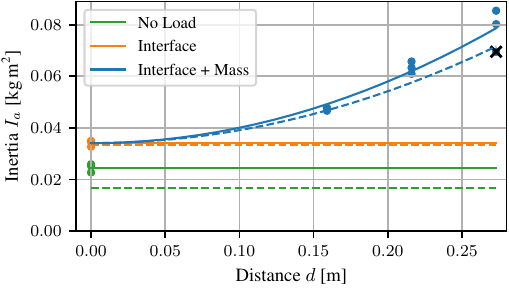}%
    \caption{Identified versus target inertia as a function of lever arm radius, with drive compensations disabled.}
    \label{fig:appendix:compansations_on}
\end{figure}

\begin{figure}
    \centering
    \includegraphics[width=\linewidth]{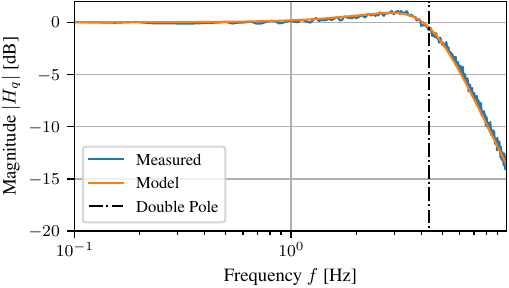}
    \caption{An example bode plot from the single-drive experiment for the closed-loop transfer function $H_q : \hat{q}_j \rightarrow q_j$, mapping commanded joint position $\hat{q}_j$ to measured joint position $q_j$.}
    \label{fig:appendix:bode_plot}
\end{figure}

\begin{figure}
    \centering
    \includegraphics[width=\linewidth]{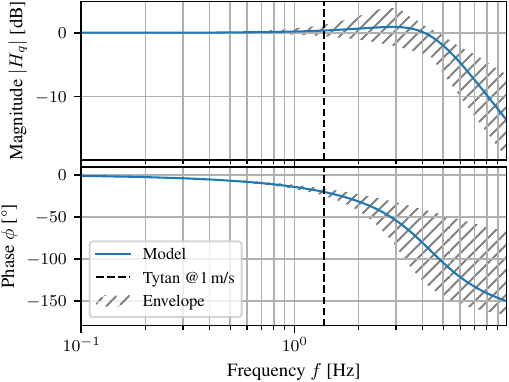}
    \caption{An example bode plot from the single-drive experiment for the closed-loop transfer function $H_q : \hat{q}_j \rightarrow q_j$, mapping commanded joint position $\hat{q}_j$ to measured joint position $q_j$. The hatched region indicates the $2\sigma$ confidence interval from analytic estimation (Appendix~\ref{sec:appendix:link_inertia_estimation}).}
    \label{fig:appendix:domain_rand}
\end{figure}

\begin{figure}
    \centering

    \begin{subfigure}{\linewidth}
        \centering
        \includegraphics[width=\linewidth]{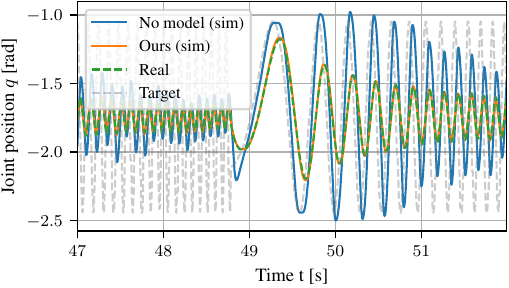}
       \caption{Evaluation on the chirp trajectory used for parameter identification.}
        \label{fig:appendix:minimal_chirp}
    \end{subfigure}

    \vspace{0.6em}

    \begin{subfigure}{\linewidth}
        \centering
        \includegraphics[width=\linewidth]{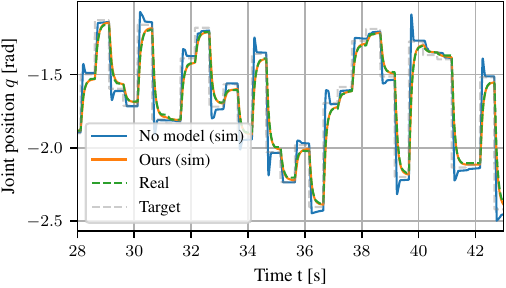}
        \caption{Validation on a random-step trajectory not used for parameter identification.}
        \label{fig:appendix:minimal_random}
    \end{subfigure}

    \caption{Trajectory tracking performance of the LF KFE joint on the \robot{Minimal} platform. The identified parameters are evaluated on (a) the chirp trajectory used for fitting and (b) an unseen random-step trajectory.}
    \label{fig:appendix:minimal_sim2real}
\end{figure}

\begin{figure*}
    \centering

    \begin{subfigure}{\linewidth}
        \centering
        \includegraphics[width=\linewidth]{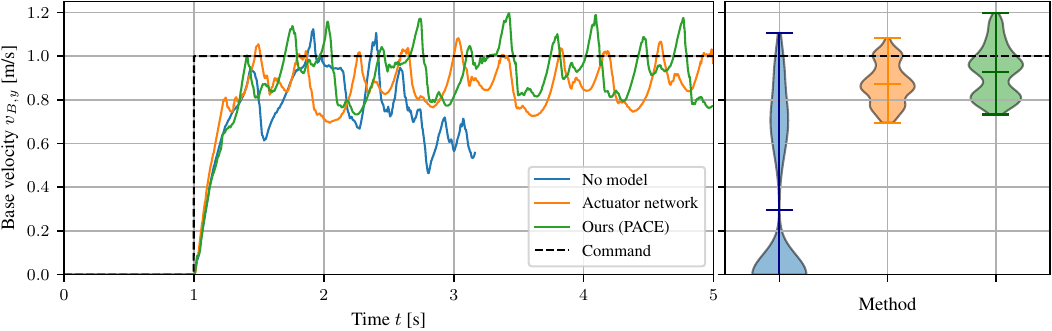}
        \caption{Sideways velocity tracking. Left: commanded and measured velocities over time. Right: distribution of steady-state velocity errors across methods.}
        \label{fig:results:anymal:sideways_velocity_plot}
    \end{subfigure}

    \medskip

    \begin{subfigure}{\linewidth}
        \centering
        \includegraphics[width=\linewidth]{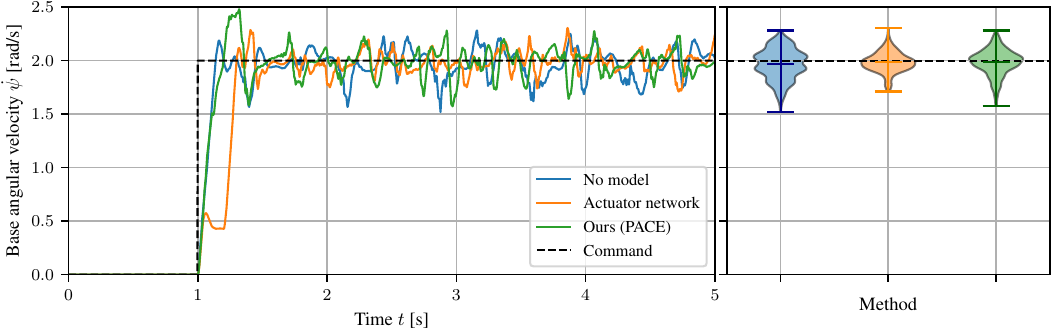}
        \caption{Angular velocity tracking around the yaw axis. Left: commanded and measured velocities over time. Right: distribution of steady-state velocity errors across methods.}
        \label{fig:results:anymal:rotation_velocity_plot}
    \end{subfigure}

    \caption{Additional sim-to-real evaluation of \robot{ANYmal}. 
    (a) Sideways velocity tracking. (b) Yaw angular velocity tracking. Both panels compare commanded and measured base velocities over time (left) and summarize steady-state errors across methods (right).}
    \label{fig:results:anymal:appendix}
\end{figure*}

\begin{figure}
    \centering
    \includegraphics[width=\linewidth]{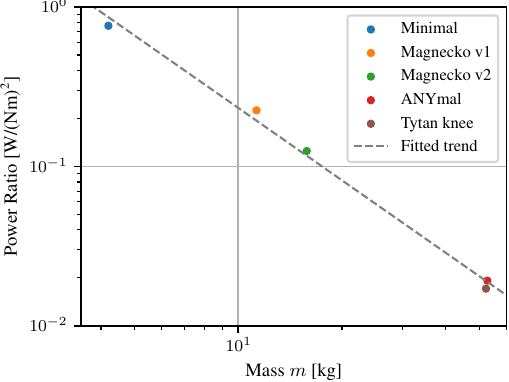}
    \caption{Joule heating constants of different robots, derived from motor parameters over body weight. These values offer a first-order estimate for black-box robotic systems.}
    \label{fig:appendix:joule_heating_constant}
\end{figure}

\section{Voltage–Limited Torque Bandwidth of a PMSM}\label{sec:appendix:pmsm_limit}

Even with perfect control, a PMSM’s current (and thus torque) loop is fundamentally limited by the available DC–bus voltage and the phase inductances. Under a low–speed / locked–rotor assumption with \(i_d\!\approx\!0\) and negligible delay, the \(q\)-axis dynamics reduce to
\begin{align}
    v_q(t) &= R\,i_q(t) + L\,\dot{i}_q(t), \\
    \Rightarrow\quad \frac{I_q(s)}{V_q(s)} &= \frac{1}{L s + R}, \label{eq:appendix:pmsm_first_order}
\end{align}
with pole at \(-R/L\), time constant \(\tau=L/R\), and theoretical \(-3\,\mathrm{dB}\) bandwidth
\begin{align}
    \omega_\infty = \frac{R}{L}, \qquad
    f_\infty = \frac{1}{2\pi}\frac{R}{L}. \label{eq:appendix:pmsm_bandwidth}
\end{align}
Taking the available phase voltage to be limited by the DC–bus (conservatively \(U_{\max}\!\approx\!U_\mathrm{bus}=\SI{48}{V}\), up to modulation factors), our drive yields \(f_\infty \approx \SI{310}{Hz}\). Hence, even at the voltage limit the current loop can, in principle, exceed the policy Nyquist of \(\SI{25}{Hz}\).

At nonzero electrical speed \(\omega_e\), the back–EMF \(e_q=\omega_e \lambda\) consumes part of the voltage headroom:
\begin{align}
    v_q = R\,i_q + L\,\dot{i}_q + e_q
    \;\;\Rightarrow\;\;
    L\,\dot{i}_q \le U_{\max} - R\,i_q - e_q.
\end{align}
As \(\omega_e\) increases (or during field–weakening), the effective current (torque) bandwidth reduces because \(U_{\max}-e_q\) shrinks. In practice, the achievable closed–loop bandwidth is also bounded to a fraction of the PWM carrier frequency and by inverter dead times and delays.

\noindent\textbf{Implication.} For our experiments—low to moderate speeds and \(\le\)\(\SI{25}{Hz}\) policy content—the inner loop behaves as a near–unit–gain motor torque source; high–speed limits are then dominated by voltage/current headroom (e.g., battery current limiting), not by inner–loop dynamics.

\section{Effective Base Inertia Comparison on ANYmal} \label{sec:appendix:effective_base_inertia}

To analyze the difference between drive-specific in-air inertia and contact-induced inertia, we apply d'Alembert's principle of virtual displacements. We consider two scenarios: (i) vertical motion of the robot’s base and legs, and (ii) horizontal motion parallel to the ground, both starting from the nominal \emph{agile stance}. In both cases, we analyze the dynamics in the robot’s $x$–$z$ plane.

For tractability, we neglect leg masses and account only for the base mass and link inertias. We further assume equal link lengths of \qty{0.3}{\meter} for the thigh and shank and no gravity.

\begin{table*}
\small\sf\centering
\caption{Dynamics properties of the \robot{Tytan} leg: CAD values, experimental estimates, and propagated uncertainties.}
\label{tab:measurements}
\begin{tabular}{c|c|c|c|c}
\toprule
& CAD & Measured & Difference \% & Measurement Error \\ 
\hline \hline
$\parallel r_1 \parallel$ [\si{\metre}] & -- & \num{0.30} & -- & \num{\pm 0.01} \\
$\parallel r_2 \parallel$ [\si{\metre}] & \num{0.24} & \num{0.25} & \num{4.2} & \num{\pm 0.01} \\
$\parallel r_3 \parallel$ [\si{\metre}] & \num{0.075} & \num{0.105} & \num{40} & \num{\pm 0.01} \\
Gravity $g$ [\si{\metre\per\second\squared}] & \num{9.81} & \num{9.806} & \num{0.0} & -- \\

Thigh Mass $m_T$ [\si{\kilogram}] & \num{3.612} & \num{3.775} & \num{4.5} & -- \\
Shank Mass $m_S$ [\si{\kilogram}] & \num{0.4535} & \num{0.4803} & \num{5.9} & -- \\
Thigh Eigenfrequency $f_T$ [\si{\hertz}] & -- & \num{0.82} & -- & \num{\pm 0.03} \\ 
Shank Eigenfrequency $f_S$ [\si{\hertz}] & -- & \num{0.88} & -- & \num{\pm 0.03} \\
\hline
Thigh Inertia $I_T$ [\si{\kilogram\metre\squared}] & \num{0.0228} & \num{0.0786} & \num{240} & \num{\pm 0.0318} \\
Shank Inertia $I_S$ [\si{\kilogram\metre\squared}] & \num{0.0107} & \num{0.00850} & \num{-21} & \num{\pm 0.00276} \\
CoM induced Thigh Inertia [\si{\kilogram\metre\squared}] & \num{0.0203} & \num{0.0416} & \num{100} & \num{\pm 0.00793} \\
\bottomrule
\end{tabular}
\end{table*}

\textbf{Vertical Motion.}  
This case is simplest, as no torque is applied at the HFE joint, and d'Alembert’s principle reduces to Eq.~\eqref{eq:appendix:dalembert_vertical}:
\begin{align}
    m \ddot{z} \delta z + I_h\ddot{q}_{h}\delta q_h + I_k\ddot{q}_{k}\delta q_k &= 2\tau_k \delta q_k \label{eq:appendix:dalembert_vertical} \\
    \text{s.t.} \quad x_B &= 0,
\end{align}
where $m$ is the base mass, and $I_h$ and $I_k$ denote the reflected and body inertias of the thigh and shank, respectively. The kinematic constraints yield the following dependencies between the knee joint angle $q_k$, the hip joint angle $q_h$, and the base $z$-position $z_B$:
\begin{align}
    q_k &= 2 q_h \label{eq:appendix:vertical_q_h} \\
    z_B &= 2l \cos\left(\tfrac{q_k}{2}\right) \label{eq:appendix:vertical_z} \\
    \delta z_B &= -l \sin\left(\tfrac{q_k}{2}\right) \delta q_k \\
    \dot{z}_B &= -l \sin\left(\tfrac{q_k}{2}\right) \dot{q}_k \\
    \ddot{z}_B &= -l \sin\left(\tfrac{q_k}{2}\right) \ddot{q}_k - \tfrac{l}{2} \cos\left(\tfrac{q_k}{2}\right)\dot{q}_k^2.
\end{align}

Substituting these relations into Eq.~\eqref{eq:appendix:dalembert_vertical} yields the nonlinear differential equation
\begin{align}
    \tau_k &= \tilde{I}_k \ddot{q}_k
    + \tfrac{1}{4}ml^2 \sin\left(\tfrac{q_k}{2}\right)\cos\left(\tfrac{q_k}{2}\right) \dot{q}_k^2 \label{eq:appendix:vertical_reduced} \\
    \tilde{I}_k &= \tfrac{1}{2}\Big[ ml^2\sin\!\left(\tfrac{q_k}{2}\right) + \tfrac{1}{4} I_h + I_k \Big].
\end{align}

\begin{figure}
    \centering
    \subfloat[\robot{ANYmal} in its agile stance with marked reference variables.]{%
        \includegraphics[width=\linewidth]{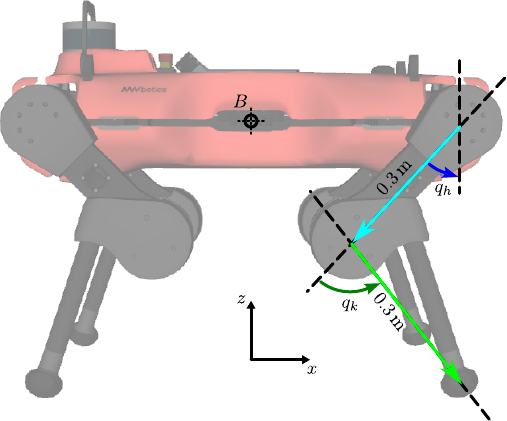}%
        \label{fig:appendix:anymal_agile_stance}}
    \vfill
    \subfloat[Effective reduced inertia at the KFE joint during vertical base motion, induced by the base's mass.]{%
        \includegraphics[width=\linewidth]{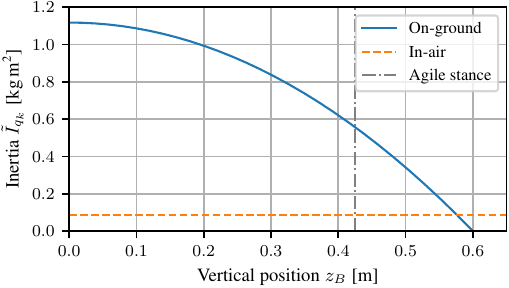}%
        \label{fig:appendix:base_inertia_vertical}}
    \vfill
    \subfloat[Effective reduced inertia at the HFE joint during horizontal base motion, induced by the base's mass.]{%
        \includegraphics[width=\linewidth]{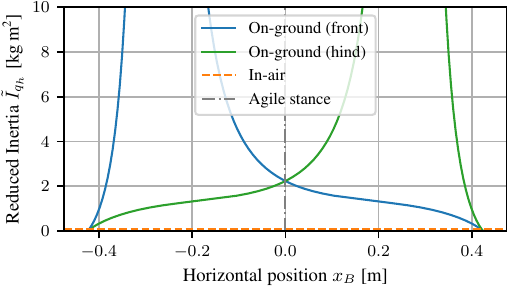}%
        \label{fig:appendix:base_inertia_horizontal}}
    \caption{Reduced inertia comparison for \robot{ANYmal} with and without ground contact. Panel (a) shows the agile stance used for evaluation. Panels (b) and (c) depict the reduced inertia at the KFE and HFE joint respectively as a function of base position during vertical and horizontal base motion, respectively.}
    \label{fig:appendix:inertia_comparison_air_ground}
\end{figure}

Assuming small velocities $\dot{q}_k \ll 1$, we plot the position–dependent inertia against the base position in Fig.~\ref{fig:appendix:base_inertia_vertical}, and compare it to the constant inertia observed when the legs are not in contact.

\textbf{Horizontal Motion.} 
The horizontal case is more involved, leading to large coupled equations. We employ a differentiable symbolic toolbox to obtain the reduced dynamics. As in the vertical case, we report only the main expressions: the virtual displacement in the $x$ direction (Eq.~\ref{eq:appendix:dalembert_horizontal}) and the base constraint (Eq.~\ref{eq:horizontal:constraint}).
\begin{align}
    m \ddot{x} \delta x + I_h\ddot{q}_{h}\delta q_h + I_k\ddot{q}_{k}\delta q_k &= 2\tau_h \delta q_h \label{eq:appendix:dalembert_horizontal} \\
    \text{s.t.} \quad z_B &= \sqrt{2}\, l \label{eq:horizontal:constraint}
\end{align}

The relations between the base $x$-position $x_B$ and the joint angles are highly nonlinear and coupled:
\begin{align}
    x_B &= l\!\left(\cos(q_k-q_h) + \cos(q_h)\right) \label{eq:appendix:horizontal_x_b} \\
    q_k &= q_h \pm \arccos\!\left( \sqrt{2} - \cos(q_h) \right). \label{eq:appendix:horizontal_q_h_k}
\end{align}
Because the dependencies are strongly nonlinear, their derivatives are large, and the effective inertia varies accordingly. The results are shown in Fig.~\ref{fig:appendix:base_inertia_horizontal}.

\begin{figure}
    \centering
    \includegraphics[width=\linewidth]{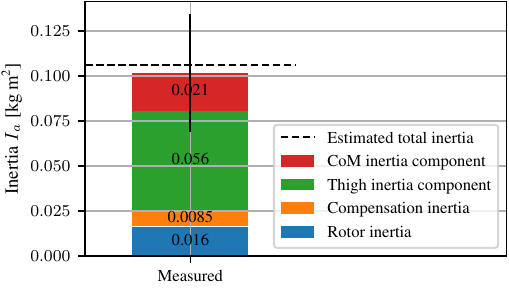}
    \caption{Breakdown of reduced inertia contributions for the full robot configuration for \robot{Tytan} in Section~\ref{sec:experiments:full_robot}. The bar plot shows the distribution of components shaping the fitted joint inertia estimate $I_a$ (dotted black line) obtained with \ac{pace}. The solid line denotes the $\pm1\sigma$ confidence interval from analytic measurements in Appendix~\ref{sec:appendix:link_inertia_estimation}.}
    \label{fig:results:single_leg_inertia}
\end{figure}

\begin{figure}
    \centering
    \includegraphics[width=\linewidth]{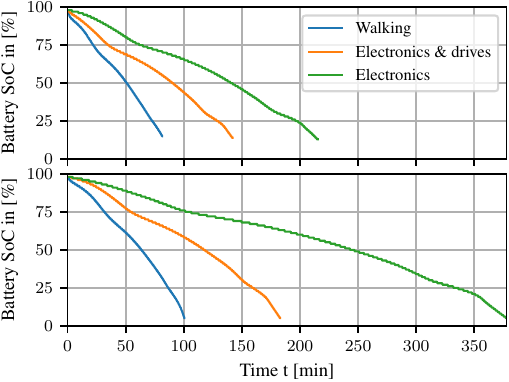}
    \caption{Full battery depletion time of \robot{ANYmal} (Top) and \robot{Tytan} (bottom) at walking in blue, resting (Electronics \& drives) in orange, and lying on the ground in green (Electronics).}
    \label{fig:appendix:times_soc}
\end{figure}

\section{Link Inertia Estimation} \label{sec:appendix:link_inertia_estimation}

To analyze the discrepancy between single-drive and full-robot inertia estimates, we performed a study on a leg-segment of \robot{Tytan}. The KFE was disconnected to minimize friction, allowing the thigh or shank to swing freely. This setup permits direct estimation of link inertias from free oscillations, which we then compare against CAD-derived values and the \ac{pace} fit.

\subsection{Pendulum Model}
Considering small oscillations around a pivot point $P$, the dynamics of the harmonic oscillator are
\begin{align}
    I_p \ddot{q} + m r g q &= 0, \label{eom}
\end{align}
with link mass $m$, CoM distance $r$, and gravitational acceleration $g$. The standard form is
\begin{align}
    \ddot{q} + \omega^2 q &= 0, \label{eom_standard_form}
\end{align}
where $\omega = 2\pi f$ is the eigenfrequency. By comparing Eq.~\eqref{eom} and Eq.~\eqref{eom_standard_form}, the inertia about the pivot $P$ is
\begin{align}
    I_P \;=\; \frac{m r g}{(2\pi f)^2}.
\end{align}
Using the parallel-axis theorem, the inertia about the CoM follows as
\begin{align}
    I_{\text{CoM}} \;=\; I_{P} - m r^2.
\end{align}

\subsection{Measured Values and Uncertainties}
We assume negligible uncertainty in $m$ and $g$, while $r$ and $f$ are measured with uncertainties of \qty{1}{\centi\metre} and \qty{0.3}{\hertz} (ten-cycle timing). Error propagation is computed via
\begin{align}
    \sigma^2 \;=\; \sum_i \left(\frac{\partial f}{\partial x_i}\right)^2 \sigma_{x_i}^2, \label{eq:error_propagation}
\end{align}
and applied to both thigh and shank. The resulting inertias and variances are summarized in Table~\ref{tab:measurements}, which also contrasts experimental and CAD values.

\subsection{Results}
For the thigh, the inertia about the CoM is
\begin{align}
    I_{T,\mathrm{CoM}} &= \SI{0.0786}{\kilogram\metre\squared}
                         \pm \SI{0.0318}{\kilogram\metre\squared},
\end{align}
while the shank yields
\begin{align}
    I_{S,\mathrm{CoM}} &= \SI{0.0085}{\kilogram\metre\squared}
                         \pm \SI{0.0028}{\kilogram\metre\squared}.
\end{align}
Figure~\ref{fig:results:single_leg_inertia} illustrates the resulting dynamics envelopes, including $2\sigma$ uncertainty bounds. Compared to CAD measurements (Table~\ref{tab:measurements}), the thigh inertia is noticeably higher, while the shank remains consistent.

These results explain part of the fourfold increase observed in the full-robot $I_a$ and confirm that the fitted inertia represents the combined effects of rotor, link, and compensation dynamics. Figure~\ref{fig:appendix:domain_rand} further visualizes the parameter spread under uncertainty, showing how these variations remain fully captured by \ac{pace}.

\begin{table}
\small\sf\centering
\caption{PPO Hyperparameters for RL Pipeline} 
\label{tab:appendix:ppo_hyperparameters}
\begin{tabular}{l|l}
\toprule
\textbf{Parameter} & \textbf{Value} \\
\hline
\hline
Empirical normalization & True \\
Number of iterations & \num{30000} \\
Value loss coefficient $c_v$ & \num{1.0} \\
Clipped value loss & True \\
Clipping parameter $\epsilon$ & \num{0.2} \\
Entropy coefficient $\alpha$ (initial) & \num{2e-3} \\
Entropy coefficient $\alpha$ (final) & \num{5e-4} \\
Entropy decay turn-over point & \num{20000} \\
Number of learning epochs $N_\text{epoch}$ & \num{5} \\
Number of mini-batches $N_\text{mb}$ & \num{10} \\
Learning rate $\eta$ & \num{e-3} \\
Learning rate schedule & adaptive \\
Discount factor $\gamma$ & \num{0.99} \\
GAE parameter $\lambda$ & \num{0.95} \\
Desired KL divergence $D_\text{KL}^\text{target}$ & \num{e-2} \\
Max gradient norm $\|\nabla\|_\text{max}$ & \num{1.0} \\
Actor hidden nodes & $[256,\,256,\,256,\,128]$ \\
Critic hidden nodes & $[256,\,256,\,256,\,128]$ \\
Initial policy std $\sigma_0$ & \num{1.5} \\
Steps per environment $N_\text{step}$ & \num{24} \\
Activation function & exponential linear unit (ELU) \\
\bottomrule
\end{tabular}
\end{table}

\end{document}